\documentclass[lettersize,journal]{IEEEtran}
\usepackage{amsmath,amsfonts}
\usepackage{algorithmic}
\usepackage{algorithm}
\usepackage{array}
\usepackage[caption=false,font=normalsize,labelfont=sf,textfont=sf]{subfig}
\usepackage{textcomp}
\usepackage[table]{xcolor}  
\usepackage{stfloats}
\usepackage{url}
\usepackage{verbatim}
\usepackage{graphicx}
\usepackage{tabularx}
\usepackage{booktabs}
\usepackage{makecell}
\usepackage{cite} 
\usepackage[colorlinks,
            linkcolor=blue,
            anchorcolor=blue,
            citecolor=blue]{hyperref}

\usepackage{doi}

\usepackage{url}
\begin{document}

\title{DCL-SE: Dynamic Curriculum Learning for \\Spatiotemporal Encoding of Brain Imaging}
\author{Meihua Zhou, 
        Xinyu Tong,
        Jiarui Zhao,
        Min Cheng,
        Li Yang,
        Lei Tian, 
        and Nan Wan
        
\thanks{Manuscript received April 19, 2021; revised August 16, 2021. %
\textit{  (Corresponding authors: Lei Tian and Nan Wan)}}
\thanks{Meihua Zhou, Xinyu Tong  and Jiarui Zhao are graduate students at the University of Chinese Academy of Sciences, Beijing 100049, China. (e-mail: mhzhou0412@gmail.com; tonxycs@gmail.com; zhaojiarui@binn.cas.cn)}%
\thanks{Min Cheng is with the Information Department, Wuhu Hospital, Beijing Anding Hospital, Capital Medical University, Wuhu, Anhui 241000, China. (e-mail: minchenginfo@163.com)}%
\thanks{Lei Tian is a Professor and Deputy Chief Physician with Beijing Tongren Hospital, Capital Medical University, and the Beijing Institute of Ophthalmology, Beijing 100730, China. He is also a postgraduate advisor. (e-mail: tianlei0131@163.com)}%
\thanks{Li Yang and Nan Wan is an Associate Professor with Wannan Medical university, Wuhu, Anhui 241002, China. (e-mail: wannan@wnmc.edu.cn; yangli@wnmc.edu.cn)}%
}


\markboth{Journal of \LaTeX\ Class Files,~Vol.~14, No.~8, August~2021}%
{Shell \MakeLowercase{\textit{et al.}}: A Sample Article Using IEEEtran.cls for IEEE Journals}


\maketitle

\begin{abstract}
High-dimensional neuroimaging analyses for clinical diagnosis are often constrained by compromises in spatiotemporal fidelity and the limited adaptability of large-scale, general-purpose models. To address these challenges, we introduce Dynamic Curriculum Learning for Spatiotemporal Encoding (DCL-SE), an end-to-end framework centered on data-driven spatiotemporal encoding (DaSE). We leverage Approximate Rank Pooling (ARP) to efficiently encode three-dimensional volumetric brain data into information-rich, two-dimensional dynamic representations, and then employ a dynamic curriculum learning strategy—guided by a Dynamic Group Mechanism (DGM)—to progressively train the decoder, refining feature extraction from global anatomical structures to fine pathological details. Evaluated across six publicly available datasets—including Alzheimer’s disease and brain tumor classification, cerebral artery segmentation, and brain age prediction—DCL-SE consistently outperforms existing methods in accuracy, robustness, and interpretability. These findings underscore the critical importance of compact, task-specific architectures in the era of large-scale pretrained networks.
\end{abstract}


\section{Introduction}
\IEEEPARstart{N}{eurological} disorders such as Alzheimer's disease (AD) and brain tumors remain critical, unsolved challenges in clinical medicine, where early and accurate diagnosis can profoundly influence patient outcomes\cite{iadecola2004neurovascular,martinez2021alzheimer,tnnlsbrain}. Modern neuroimaging modalities—especially Magnetic Resonance Imaging (MRI) and Computed Tomography (CT)—enable detailed, noninvasive visualization of brain structure\cite{hatami2024investigating,menagadevi2024machine,wang2022high}. However, translating such high-dimensional imaging data into robust, actionable diagnostic decisions remains challenging due to subtle clinical signals, spatial heterogeneity, and limited annotated datasets\cite{roy2015thr}.

A common strategy to alleviate computational burdens in neuroimaging is to project 3D brain volumes into 2D slices for analysis \cite{soomro2023ima,liu2024mult,zhang2023tw}. While computationally efficient, this approach incurs a critical trade-off: the loss of spatial continuity and essential anatomical relationships within the brain. Such simplification disrupts global spatial coherence, complicating the capture of critical pathological patterns, especially when lesions or atrophic regions are sparse or irregularly distributed. Moreover, transferring information from 3D to 2D exacerbates the challenge by losing both spatial and semantic context, which are crucial for accurate disease characterization \cite{woo2021com,tnnls3d}. This issue is further compounded by the limitations of traditional learning strategies, such as Curriculum Learning (CL). These methods often rely on static difficulty levels, which fail to adapt to the complexities inherent in neuroimaging data. Conventional CL methods use manually defined stages that do not dynamically adjust to varying feature complexities, thus hindering the model’s ability to progressively learn from simple to complex features. As both spatial relationships and pathologies evolve throughout the learning process, this static approach leads to models failing to disentangle fine-grained features, resulting in suboptimal performance. Consequently, there is a clear need for a new approach—one that not only bridges the cross-modal gap between 3D and 2D but also incorporates an adaptive, dynamic learning process that progressively captures and refines diagnostically relevant features (See the related work section below for further details.).

\begin{figure}[htbp]
    \centering
    \includegraphics[width=0.45\textwidth]{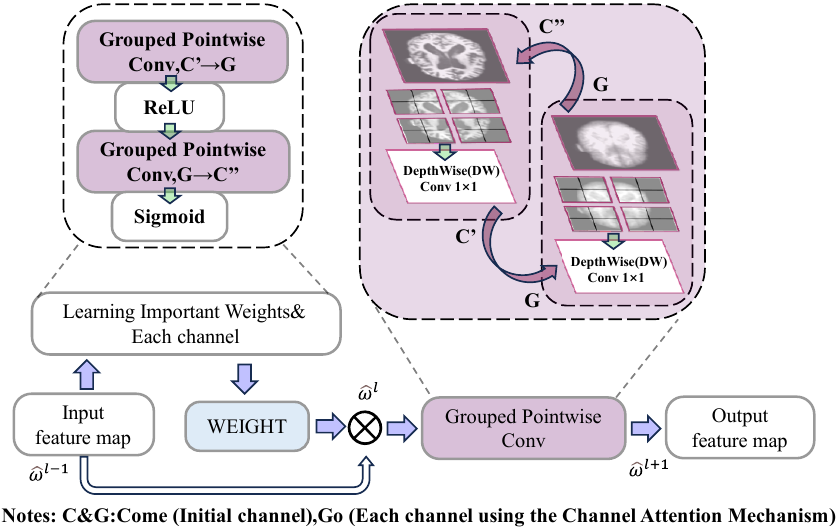}
    \caption{{\small Dynamic Grouping Mechanism (DGM), an important component of DCL-SE and the carrier of Dynamic Curriculum Learning (DCL), centralizes DCL by generating, weighting, and adjusting dynamic features. These features are then processed through grouped convolution to extract critical channel-wise information.}}
    \label{fig:DGM}
\end{figure}

Recent advances in large-scale neural network models have promoted the development of universal, transferable architectures, including vision and multimodal frameworks such as MedSAM\cite{medsam}, BioGPT\cite{biogpt}, and MedCLIP\cite{medclip}, demonstrating promising cross-domain generalization and zero-shot learning capabilities\cite{medsam3d}. Nevertheless, the practical utility of these large models in neuroimaging remains constrained by data scarcity, stringent privacy requirements, and significant computational costs. In our systematic evaluation of eleven state-of-the-art large models (GPT-4\cite{gpt4}, Gemini\cite{gemini}, Claude\cite{claude}, Pixtral\cite{pixtral}, ERNIE\cite{ernie}, Flux\cite{flux}, GLM\cite{glm4}, HunYuan-lite\cite{hunyyuanlite}, Moonshot-V1\cite{moonshotv1vision}, among others) on Alzheimer's disease MRI tasks via direct zero-shot inference, we find these off-the-shelf architectures consistently underperform specialized domain-specific frameworks—even with advanced prompt engineering. This observation reveals a critical gap between theoretical generalization advances and practical application constraints, motivating tailored lightweight solutions designed explicitly for clinical neuroimaging tasks.

To address these pressing challenges, we introduce Dynamic Curriculum Learning for Spatiotemporal Encoding (DCL-SE), an end-to-end lightweight framework specifically developed for robust and interpretable brain imaging analysis. Central to DCL-SE is the concept of Data-based Spatiotemporal Encoding (DaSE), which comprises two sequential stages: an encoding stage and a decoding stage. In the encoding stage, DaSE employs Approximate Rank Pooling (ARP) to efficiently distill three-dimensional volumetric brain data into compact yet information-rich two-dimensional dynamic representations, thus preserving crucial spatial and anatomical progression contexts essential for accurate clinical diagnosis. However, directly training on these ARP projections may lead to entangled representations and obscure subtle pathological cues. To address this issue, the decoding stage of DaSE adopts a dynamic curriculum learning strategy guided by a Dynamic Group Mechanism (DGM), which progressively disentangles and refines diagnostically salient features from the encoded representations. As illustrated in Figure~\ref{fig:DGM}, DGM adaptively recalibrates feature importance, mimicking the hierarchical reasoning process of clinical experts—initially focusing on global anatomical structures and subsequently refining attention toward subtle pathological details.

Beyond disease classification, we demonstrate the versatility of DCL-SE across multiple neuroimaging tasks, including brain tissue segmentation and age prediction, accompanied by comprehensive interpretability analyses via t-SNE projections, heatmaps, and saliency mapping. Extensive experiments on diverse public datasets confirm that DCL-SE consistently surpasses existing conventional architectures and recent large-scale pretrained models in accuracy, efficiency, and interpretability.

The main contributions of this work are summarized as follows:

1) We develop the DCL-SE framework, a lightweight adaptive model designed to dynamically prioritize feature extraction according to complexity and clinical significance.

2) Proposal of DaSE, a data-driven spatiotemporal encoding method composed of two sequential stages: encoding via ARP to transform volumetric brain data into compact dynamic representations, and decoding via Dynamic Curriculum Learning guided by the DGM.

3) Comprehensive validation of DCL-SE's robustness and interpretability across representative neuroimaging tasks—including brain disease classification, cerebral artery segmentation, and brain age prediction—providing extensive visual evidence and analyses supporting its applicability.

4) A systematic empirical evaluation revealing the significant performance limitations of large general-purpose models in practical clinical neuroimaging scenarios characterized by limited samples and stringent privacy constraints.

\begin{figure*}[htbp]
    \centering
    \includegraphics[width=\textwidth]{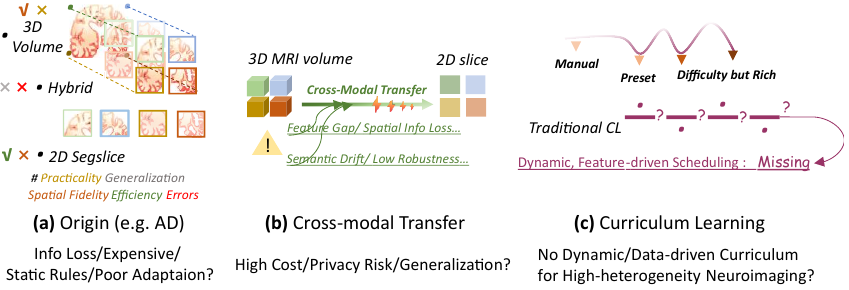}
    \caption{{\small Illustration of existing multimodal image data processing and learning strategies. (a) Static 2D, 3D, and hybrid methods are widely used in brain imaging analysis but each has limitations: 2D methods are highly efficient yet lack spatial information\cite{woo2021com, soomro2023ima}. 3D methods offer strong spatial expression but are constrained by computing power and sample size, limiting clinical practicality\cite{nie2019ima, gtifa2024mult}. hybrid methods integrate 2D and 3D via static fusion rules or post-processing but are prone to interpolation errors and poor generalizability\cite{gtifa2023com, roy2015thr}. (b) Cross-modal transfer, including 3D to 2D and CT to MRI conversions, suffers from core issues of spatial and semantic information loss during transfer\cite{tf1, tf2, kora2022transfer}. While large/foundational models enable such transfer, they face bottlenecks of high computational demands and privacy risks, with their practical generalization and information retention remaining unresolved\cite{biogpt, medclip}. (c) Curriculum Learning (CL) has notable limitations: most applications rely on static, manually defined difficulty classification, lacking feature-driven dynamic scheduling\cite{hatami2024investigating, lin2019seg, zhang2023tw}. In heterogeneous brain imaging scenarios specifically, the absence of automatic complexity measurement and adaptation mechanisms hinders gradual improvement of diagnostic complexity and generalization\cite{liu2024mult, tang2018muti}.}}
    \label{fig:Intro}
\end{figure*}

\section{Related Work}
\subsection{2D, 3D, and Hybrid Approaches for Brain Image Analysis}
Brain image analysis has advanced along three major lines: 2D slice-based methods, 3D volumetric models, and hybrid 2D-3D approaches\cite{huang2024acc,lin2019seg,tang2018muti,nie2019ima}.  
2D methods are widely adopted due to computational efficiency but fail to capture global anatomical context, leading to limited performance in complex tasks\cite{chenw2018hum}.  
3D models offer improved spatial representation and have achieved success in segmentation and disease detection\cite{tnnlsseg0,tnnlsseg}, but their high computational and data requirements restrict clinical utility.  
Hybrid approaches—such as multi-view fusion and 2.5D methods that aggregate orthogonal 2D slices—aim to balance computational efficiency with spatial awareness by approximating 3D context through coordinated 2D representations\cite{vidhya2019systematic,riyazudeen2022converting}. Some methods also incorporate interpolation to reconstruct pseudo-3D volumes from sparse slices. However, these strategies often rely on static fusion rules or shallow aggregation schemes, which may lead to loss of spatial fidelity, introduce interpolation artifacts, or fail to capture heterogeneity in small-sample neuroimaging datasets. Therefore, designing models that efficiently preserve diagnostically critical spatial information remains a key challenge in brain image analysis..  

\subsection{Transfer Learning and Foundation Models in Medical Imaging}
Transfer learning has become a cornerstone of medical image analysis, particularly under the constraints of limited labeled data and stringent privacy requirements\cite{tf1,tf2,tf3}. Conventional approaches are most effective when applied within the same modality and dimensionality—such as 2D-to-2D or 3D-to-3D tasks—where feature representations are naturally aligned and semantic gaps are minimized. In contrast, cross-modal transfer learning, such as 3D-to-2D adaptation or transferring between imaging modalities (e.g., CT to MRI), remains challenging due to substantial differences in spatial context, resolution, and underlying data distributions. Direct projection or feature transfer from 3D volumes to 2D slices often results in significant information loss, limiting the model’s capacity to generalize to clinically relevant tasks\cite{kora2022transfer}.

To mitigate these limitations, a variety of advanced strategies have been proposed. Spiral-spinning “2.75D” representations, for instance, attempt to encode volumetric structure within 2D feature spaces, aiming to retain spatial information while reducing computational cost\cite{wang20232}. Meanwhile, self-supervised learning and hybrid initialization techniques have been introduced to enhance feature robustness and address overfitting in low-resource settings\cite{tcsvtmif,wen2020effective}. Nevertheless, excessive similarity between source and target datasets can cause “inbreeding effects,” where iterative reuse of datasets degrades the benefits of transfer learning and impairs model generalization.

More recently, the emergence of foundation models—including MedSAM\cite{medsam}, pre-trained backbones for medical imaging. Trained on large-scale, heterogeneous datasets, these models offer strong potential for adaptation to downstream clinical tasks via fine-tuning. Vision-language and multimodal models, such as BioGPT\cite{biogpt}, MedCLIP\cite{medclip}, and DINOv2\cite{dinov}, further extend this paradigm by incorporating cross-domain knowledge. However, the practical deployment of such models is limited by substantial computational requirements, the scarcity of large annotated clinical datasets, and persistent privacy concerns. Critically, their efficacy in cross-modal and cross-dimensional transfer scenarios—such as 3D-to-2D adaptation or integration across different imaging modalities—remains an open problem, particularly regarding information retention, semantic alignment, and computational efficiency.

\subsection{Curriculum Learning}
Curriculum learning (CL) organizes training from easy to hard examples, and has shown effectiveness in computer vision by improving convergence and generalization\cite{cl1,cl2,cl3,cl4}. In medical imaging, existing CL strategies are rare and mostly use static or heuristic difficulty criteria, such as image blurring or staged sample complexity\cite{Burduja2021Unsupervised}. However, these methods generally lack principled, data-driven metrics for task or feature complexity, and do not adaptively adjust curricula during training.

This limitation is especially apparent in brain imaging, where high heterogeneity and diverse tasks make static curricula insufficient. To date, automatic quantification of sample complexity and dynamic scheduling remain largely unexplored in medical image analysis, particularly for complex scenarios such as subject-level brain image classification, segmentation, and age prediction.  

\section{Methodology}
\subsection{Framework Overview}

\begin{figure}[htbp]
  \centering
  \subfloat[]{  
      \includegraphics[width=0.45\textwidth]{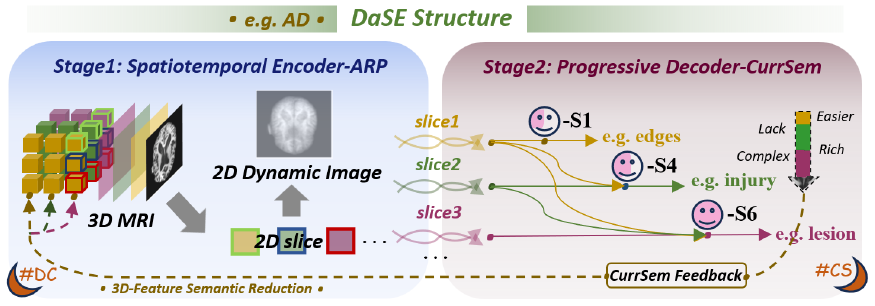}
      \label{fig:DaSE}
  } \\   
  \subfloat[]{  
      \includegraphics[width=0.45\textwidth]{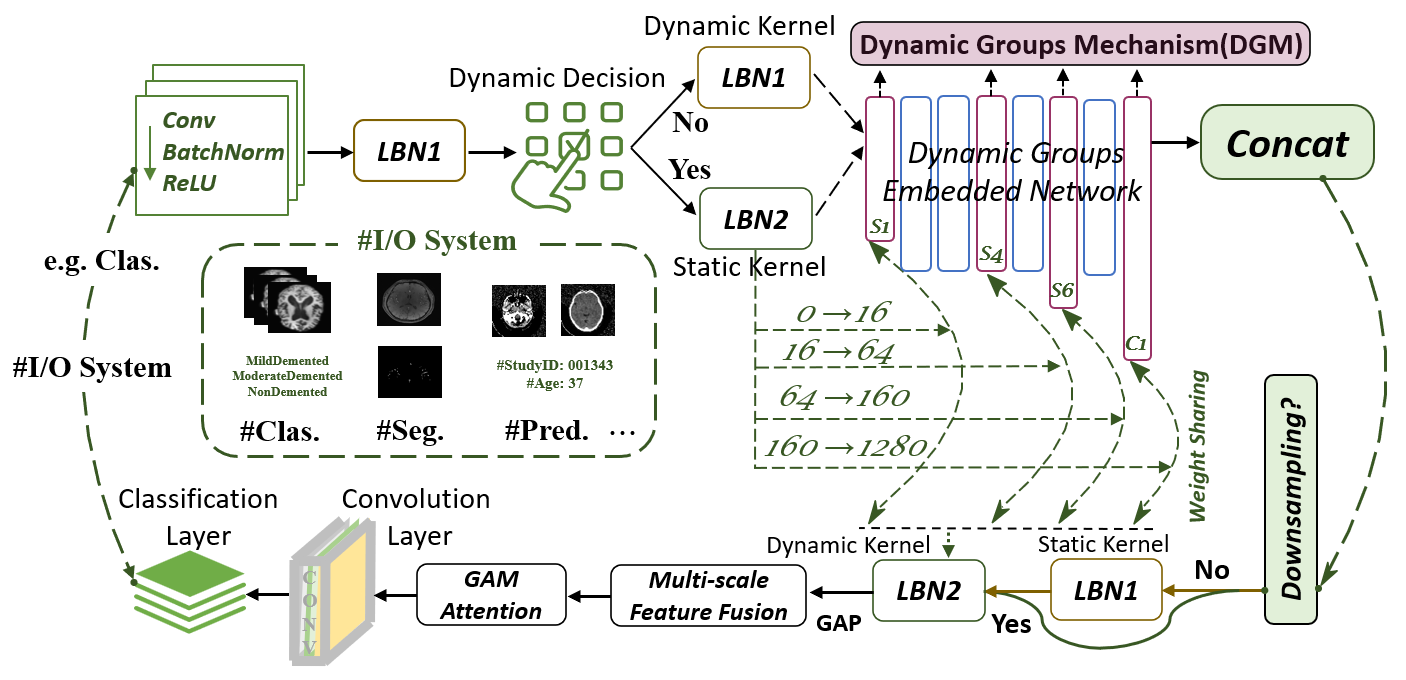}
      \label{fig:super}
  }\\
  \subfloat[]{  
      \includegraphics[width=0.45\textwidth]{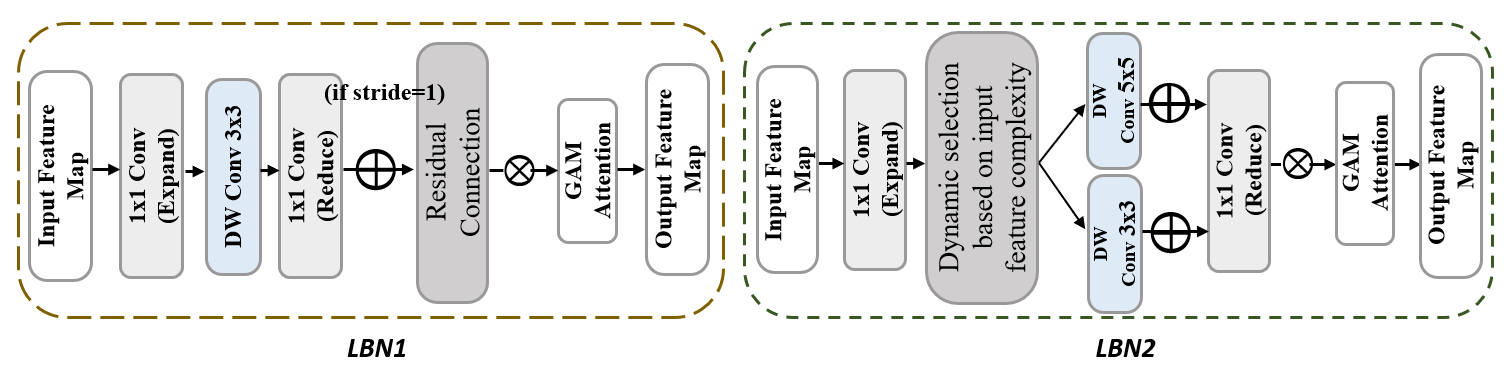}
      \label{fig:DCL-SE2}
  }
  \caption{(a) Data-based Spatiotemporal Encoding (DaSE) structure, which consists of two stages: ARP-based Spatiotemporal Encoding and Curriculum Semantic-based Progressive Decoding. For example, in Alzheimer's disease (AD), the former is a Data Conversion (DC) process that first converts the 3D MRI data into 2D slices via ARP to generate 2D dynamic images, which are used as inputs for the latter Curriculum Semantic (CurrSem) phase. At the CurrSem (CS) process, these 2D slices are then trained to decode multidimensionally through the Curriculum Learning strategy. Finally, the trained features are fed back to the 3D information via CurrSem to greatly reduce information loss between medical mltimodal data under the condition of effective training of the 3D data features. (b) DCL-SE architecture, featuring a distinctive "D" shape that represents its Dynamic Adjustment capabilities. The overall model design provides theoretical support for the decoding stage of DaSE. Specifically, the dynamic grouping mechanism (DGM) integrates the DCL strategy to perform progressive hierarchical feature extraction and, through enhanced depth separable convolution and linearization processing, improves training efficiency while reducing model size and computational complexity. (c) DCL-SE adaptively selects modules (LinearBottleNeck1 or LinearBottleNeck2) based on network positions (e.g., convolution grouping, downsampling layer placement).}
  \label{fig:ALL}
\vspace{-2.0em}
\end{figure}

Accurate subject-level analysis of brain images is fundamentally challenged by limited labeled data, high anatomical heterogeneity, and the need to balance computational efficiency with preservation of 3D spatial context. Conventional 2D or 3D models typically either sacrifice global anatomical information or incur prohibitive computational costs, while static learning pipelines do not reflect the progressive and hierarchical nature of clinical diagnostic reasoning.

To address these limitations, (See Figure~\ref{fig:super}) we propose Dynamic Curriculum Learning for Spatiotemporal Encoding (DCL-SE), a unified framework designed for efficient and interpretable analysis of heterogeneous, data-limited brain imaging scenarios. (See Figure~\ref{fig:DaSE}) The core of DCL-SE is the Data-based Spatiotemporal Encoding (DaSE) paradigm, which serves as the main architectural pipeline throughout the framework.

DaSE is composed of two tightly connected stages. In the first stage, Approximate Rank Pooling (ARP) is employed to encode the ordered spatial and temporal progression of 3D brain MRI data into a compact two-dimensional dynamic representation. This encoding step preserves diagnostically relevant spatial structure and progression cues with minimal information loss. In the second stage, the ARP-encoded dynamic image is decoded through a dynamic curriculum learning (DCL) strategy, which is implemented using a Dynamic Group Mechanism (DGM). This decoding stage progressively extracts, disentangles, and refines features from coarse anatomical context to fine pathological details, with DGM adaptively recalibrating feature importance at each step based on data complexity.

The tightly introduce encoder-decoder structure of DaSE allows DCL-SE to bridge the gap between computational efficiency and spatial fidelity. By first compressing 3D volumetric data and then decoding it in a staged and adaptive manner, the framework enables standard 2D neural networks to robustly capture and utilize the rich diagnostic information present in three-dimensional brain images. This approach is widely applicable to a variety of neuroimaging tasks, including disease classification, tissue segmentation, and regression-based predictions such as brain age.

\subsection{Data-based Spatiotemporal Encoding}
This section explores the scientific rationale underlying our approach, with a focus on the unified encoding-decoding framework that forms the foundation of our design. At the heart of this methodology is the Data-based Spatiotemporal Encoding (DaSE) paradigm, which is essential for preserving the integrity of both spatial and temporal information throughout the entire process. DaSE acts as the cornerstone of the framework, offering an efficient mechanism to encode 3D brain MRI data into 2D representations, all while retaining critical anatomical and pathological features. In the following two sections, we provide a detailed breakdown of the overall architecture and technical components of our framework, including Approximate Rank Pooling (ARP), Dynamic Curriculum Learning (DCL), and the Dynamic Group Mechanism (DGM).


A persistent challenge in brain MRI analysis is how to maintain the integrity of three-dimensional anatomical and pathological information while ensuring that the model remains practical and interpretable. In clinical settings, most efficient neural network frameworks and established diagnostic workflows rely on two-dimensional representations. However, direct 2D analysis inevitably loses a significant amount of contextual detail that defines disease progression and spatial relationships within the brain. Additionally, even when data is processed with minimal semantic loss, the training process often fails to achieve precise matching, resulting in low training efficiency despite improved data processing interventions. Consequently, the model's outcomes fall short of expectations.

To bridge this gap, our DaSE structure (See Figure \ref{fig:DaSE}) is built around a data-driven spatiotemporal encoding-decoding strategy that closely reflects the underlying scientific questions in medical imaging  (See Algorithm \ref{alg:spatio-temporal-encoding}). We view it as a critical stage that must be designed to minimize the loss of meaningful structure and dynamics.

\begin{algorithm}[!h]
    \caption{Data-based Spatiotemporal Encoding }
    \label{alg:spatio-temporal-encoding}
    \renewcommand{\algorithmicrequire}{\textbf{Input:}}
    \renewcommand{\algorithmicensure}{\textbf{Output:}}
    \begin{algorithmic}[1]
        \REQUIRE Ordered 3D MRI slices $\{\mathbf{I}_1, ..., \mathbf{I}_T\}$
        \ENSURE Task-specific prediction (e.g., brain image classification, brain image segmentation, brain age prediction)

        \STATE // Step 1: Spatiotemporal Encoding (ARP)
        \FOR{$t = 1$ to $T$}
            \STATE Extract feature vector $\psi_t \gets f_\mathrm{enc}(\mathbf{I}_t)$
        \ENDFOR
        \STATE // Step 1.2: Aggregate slice features with ARP to obtain the dynamic 2D descriptor
        \STATE $\mathbf{D}^* \gets \mathrm{ARP}(\{\psi_1, ..., \psi_T\})$
        \STATE // Step 2: Progressive Decoding (DCL with DGM)
        \STATE Initialize decoded feature $\mathbf{D}_0 \gets \mathbf{D}^*$
        \FOR{stage $i = 1$ to $N$}
            \STATE Stage-wise feature extraction and complexity evaluation
            \STATE $\mathbf{D}_i \gets \mathrm{DCLStage}(\mathbf{D}_{i-1})$
            \STATE Compute complexity $\lambda(\mathbf{D}_i)$
            \IF{$\lambda(\mathbf{D}_i) > \tau_i$}
                \STATE Apply dynamic group mechanism for recalibration
                \STATE $\mathbf{D}_i \gets \mathrm{DGM}(\mathbf{D}_i)$
                \STATE Proceed to next stage
            \ELSE
                \STATE Continue current stage refinement
            \ENDIF
        \ENDFOR

        \STATE // Step 3: Final Prediction
        \STATE $\hat{y} \gets \mathrm{Predict}(\mathbf{D}_N)$
        \RETURN $\hat{y}$
    \end{algorithmic}
\end{algorithm}

Our method begins with an encoding process: Approximate Rank Pooling (ARP) projects the ordered stack of 3D MRI slices into a single 2D dynamic image, designed to retain both the spatial structure and the progression of anatomical changes across the volume. This ensures that diagnostically relevant temporal and spatial patterns are not discarded, but instead made accessible to subsequent 2D processing.

However, the complexity of real brain data means that these compressed dynamic images still contain highly entangled features. Important pathological cues, subtle tissue boundaries, and global context may overlap in ways that challenge direct interpretation by standard convolutional networks. Recognizing this, we introduce a decoding process based on Dynamic Curriculum Learning (DCL), supported by a Dynamic Group Mechanism (DGM). This stepwise decoding does not simply “unpack” the compressed image, but instead guides the model to progressively disentangle and interpret features from coarse, easily recognizable structures to fine-grained pathological signatures, much as a clinician would approach a complex case.

By explicitly combining the encoding and decoding stages, our framework not only reduces conversion losses in multimodal data, promotes robust hierarchical learning and model transparency, but also flexibly retains and transmits the semantic information of multidimensional features after training through Curriculum Semantic Feedback (CurrSem Feedback), thereby maximizing the positive feedback regulation effect throughout the entire data processing and training process. This method, which we refer to as Data-driven Spatio-temporal Encoding (DaSE), enables actual 2D neural networks to fully leverage the available information depth in multi-dimensional brain MRI, ultimately supporting accurate, interpretable, and clinically meaningful analysis.

\subsection{Dynamic Image Construction and Cross-Modal Transfer via Approximate Rank Pooling}

Analysis of medical images often involves bridging substantial gaps between different data organizations and modalities. In conventional medical imaging research, “cross-modal” generally refers to the transfer of knowledge between different imaging mechanisms such as CT and MRI, or between various MRI sequences (e.g., T1- and T2-weighted images). However, a significant and often overlooked modality gap also exists between 2D MRI—typically stored as individual DICOM slices—and 3D volumetric MRI, such as NIfTI (.nii) data, which contain richer spatial context and different organizational structures. Beyond modality, there is also a dimensionality gap: models designed for 2D images are not directly compatible with 3D data due to mismatches in spatial continuity and feature distribution.

Our approach addresses both the cross-dimensional and cross-modal transfer challenges inherent in brain imaging analysis. The core scientific problem is how to exploit robust, well-generalized feature representations learned from abundant 2D annotated data for efficient and accurate analysis of volumetric 3D MRI, while preserving critical spatial and sequential (structural) information. Traditional pooling strategies, such as mean or max pooling across slices, are order-invariant and tend to obscure essential anatomical progressions and dynamic relationships, limiting their effectiveness for downstream volumetric tasks.

To overcome these challenges, we propose to summarize the spatial evolution of 3D MRI data using approximate rank pooling (ARP), a principled, order-sensitive aggregation technique originally developed for temporal video modeling\cite{arp1,arp2}. ARP enables the conversion of an ordered sequence of slices into a single “dynamic image”—a compact descriptor that preserves both appearance and spatial progression information. This makes it possible to leverage standard 2D CNN architectures and pretrained weights for volumetric medical data, thus bridging both the dimensional and modality gap efficiently.

Given a 3D MRI volume as an ordered sequence of $T$ slices $\{\mathbf{I}_1, \ldots, \mathbf{I}_T\}$, each slice is mapped to a feature vector $\psi_t \in \mathbb{R}^d$. The task is to construct a summary descriptor $\mathbf{d}^*$ that captures the structural progression along the anatomical axis. Rather than simply averaging features, ARP models the sequence as a ranking problem. The idea is to learn a function that assigns higher scores to later slices, capturing the directional change along the volume.

The optimization problem for rank pooling is defined as:
\begin{equation}
\min_{\mathbf{d}} \; \frac{\lambda}{2}\|\mathbf{d}\|^2 + \frac{2}{T(T-1)} \sum_{q>t} \max\{0, 1 - S(q|\mathbf{d}) + S(t|\mathbf{d})\}
\end{equation}
where $S(t|\mathbf{d}) = \langle \mathbf{d}, \mathbf{V}_t \rangle$ and $\mathbf{V}_t = \frac{1}{t} \sum_{i=1}^t \psi_i$ is the average of features up to slice $t$. This formulation, inspired by the RankSVM approach, encourages the summary vector $\mathbf{d}$ to reflect the ordered progression through the volume.

Direct optimization can be computationally intensive and not amenable to end-to-end training. Approximate rank pooling provides an efficient solution by taking a single gradient step from the origin, resulting in:
\begin{equation}
\mathbf{d}^* \propto \sum_{q>t} (\mathbf{V}_q - \mathbf{V}_t)
\end{equation}
This can be further expanded as:
\begin{equation}
\mathbf{d}^* \propto \sum_{q>t} \left[ \frac{1}{q} \sum_{i=1}^q \psi_i - \frac{1}{t} \sum_{j=1}^t \psi_j \right]
\end{equation}
which is equivalent to a weighted sum:
\begin{equation}
\mathbf{d}^* = \sum_{t=1}^T \alpha_t \psi_t
\end{equation}
where the coefficients $\alpha_t$ are analytically defined as:
\begin{equation}
\alpha_t = 2(T-t+1) - (T+1)(H_T - H_{t-1})
\end{equation}
and $H_t = \sum_{i=1}^t 1/i$ is the $t$-th harmonic number ($H_0=0$).

Here, $\mathbf{d}^*$ serves as a 2D dynamic descriptor summarizing the 3D volume; $\mathbf{V}_q$ and $\mathbf{V}_t$ are the mean feature vectors up to slices $q$ and $t$; and $\psi_i$ represents the feature at index $i$. This order-sensitive aggregation preserves both the spatial structure and sequential relationships of the original data, enabling direct compatibility with conventional 2D convolutional architectures. The resulting dimensionality alignment allows the weights pretrained on 2D data to be transferred efficiently to volumetric tasks, achieving both computational speed and accurate structural modeling.

The process is detailed in the following Algorithm \ref{alg:arp} :

\begin{algorithm}[!h]
    \caption{Approximate Rank Pooling}
    \label{alg:arp}
    \renewcommand{\algorithmicrequire}{\textbf{Input:}}
    \renewcommand{\algorithmicensure}{\textbf{Output:}}
    \begin{algorithmic}[1]
        \REQUIRE Ordered slice features $\{\psi_1, \ldots, \psi_T\}$
        \ENSURE Dynamic descriptor $\mathbf{d}^*$
        
        \STATE Initialize harmonic numbers: $H_0 \gets 0$
        \FOR{$t = 1$ to $T$}
            \STATE $H_t \gets H_{t-1} + 1/t$ \hfill // Compute harmonic numbers
        \ENDFOR
        \STATE Initialize descriptor: $\mathbf{d}^* \gets \mathbf{0}$
        \FOR{$t = 1$ to $T$}
            \STATE $\alpha_t \gets 2(T - t + 1) - (T + 1)(H_T - H_{t-1})$ \hfill 
            \STATE $\mathbf{d}^* \gets \mathbf{d}^* + \alpha_t \psi_t$ \hfill // Weighted aggregation
        \ENDFOR
        \RETURN $\mathbf{d}^*$
    \end{algorithmic}
\end{algorithm}

The ARP operator thus generates a compact dynamic representation from volumetric data that is compatible with standard 2D convolutional architectures and preserves essential spatial and sequential information for subsequent analysis. This compressed dynamic image serves as the input to our staged learning process, where the network progressively decodes the embedded spatiotemporal features through dynamic curriculum learning, as detailed in Section III-D.

\subsection{ Dynamic Curriculum Learning and Dynamic Group Mechanism}
In medical brain imaging, accurate diagnosis relies on a progressive understanding of anatomical structures and pathological features, from broad tissue types to subtle disease-specific cues. Motivated by this, we introduce Dynamic Curriculum Learning (DCL), which mimics the clinical reasoning process by gradually increasing the difficulty of the model's learning objectives. In DCL, each model stage (e.g., S1, S4, S6, Conv1) is associated with a different level of receptive field and anatomical abstraction. For instance, early stages (S1, S4) primarily capture global tissue structures such as gray/white matter and ventricles, while later stages (S6, Conv1) progressively capture fine boundaries, lesions, and focal pathology. The transitions between stages are governed by the complexity of learned feature maps, ensuring that the model only advances once it has sufficiently captured information at the current scale.

This approach is central to our proposed DCL-SE framework, which aims to decode the compressed spatiotemporal information from ARP projections in a stepwise manner that mimics clinical diagnostic logic. By progressively guiding the network from global anatomical context to local pathological details, DCL-SE directly addresses the scientific challenge of disentangling and reconstructing multi-scale 3D features from a compact 2D representation.

The complexity of a feature map at stage $i$ is quantitatively defined as
\begin{equation}
\lambda(X_i) = \sum_{c=1}^{C} \left| \nabla X_i^{(c)} \right|
\end{equation}
where $X_i \in \mathbb{R}^{C \times H \times W}$, $C$ is the number of channels, and $\nabla X_i^{(c)}$ denotes the spatial gradient in channel $c$. This metric reflects spatial richness and structural heterogeneity, which, in brain MRI, correspond to anatomical boundaries or regions of abnormal intensity. When $\lambda(X_i)$ exceeds a threshold $\tau_k$ (determined empirically or by the distribution of training complexities), the model transitions to the next curriculum stage. This mechanism enforces an adaptive, data-driven curriculum, rather than a static training order.

\begin{figure}[htbp]
    \centering
    \includegraphics[width=0.45\textwidth]{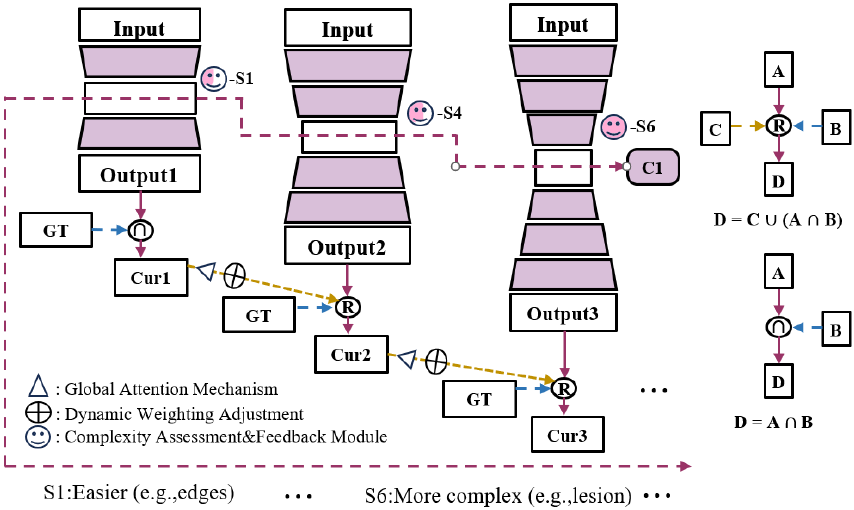}
    \caption{{\small Dynamic Curriculum Learning (DCL) runs through the whole process of S1,S4,S6,C1, the feature learning process is from simple to complex, and the complexity is evaluated and feedback is given at each stage. Feature fusion on (Cur.) intersection and merger sets, weighted merging of features from different stages to form a multidimensional important and comprehensive information representation through global attention approach.}}
    \label{fig:DCL}
\end{figure}

To enable stage-wise adaptive learning, we introduce the Dynamic Group Mechanism (DGM), which recalibrates feature importance at each stage. DGM is defined by the following computational principle:
\begin{equation}
\left\{
\begin{aligned}
\hat{w}^l &= LIW \left( \text{Sum}(\hat{w}^{l-1}) \right) \times \hat{w}^{l-1} \\
\hat{w}^{l+1} &= GPConv \left( \hat{w}^l \right)
\end{aligned}
\right.
\end{equation}
Here, $LIW(\cdot)$ (Learned Importance Weighting) denotes an adaptive weighting computed from the aggregated feature response of the previous stage, operationalized as a grouped convolution followed by a non-linearity and a sigmoid normalization, as in:
\begin{equation}
\left\{
\begin{aligned}
W &= \sigma\left( \text{Conv}_G\left( \mathrm{ReLU}\left( \text{Conv}_G(X) \right) \right) \right) \\
X' &= X \odot W \\
Y &= \text{Conv}_G(X')
\end{aligned}
\right.
\end{equation}
where $\text{Conv}_G$ is the grouped convolution, $W$ is the adaptive weight map highlighting diagnostically relevant features, $X'$ is the reweighted feature, and $Y$ is the stage output. The DGM thereby ensures that the features with high clinical relevance (such as edges, boundaries, or lesions) are adaptively prioritized at each curriculum stage.

Figure~\ref{fig:DCL} illustrates how DCL runs across S1, S4, S6, and Conv1, with the feature complexity guiding the stage transitions, and feature fusion across curriculum stages yielding a comprehensive information representation. At each step, DGM adapts channel and spatial attention, resulting in efficient feature learning and enhanced diagnostic focus.

For efficiency, DGM employs grouped convolution, which reduces the parameter count compared to standard convolution. Specifically, for input channels $C_{in}$, output channels $C_{out}$, kernel size $K$, and $G$ groups, the parameter count is
\begin{equation}
\text{Params}_{GroupConv} = \frac{C_{in} \cdot C_{out} \cdot K^2}{G}
\end{equation}
Setting $G=4$ for example, the parameter count is reduced by a factor of 4, making DGM suitable for data-limited medical imaging scenarios.

\begin{figure}[htbp]
    \centering
    \includegraphics[width=0.45\textwidth]{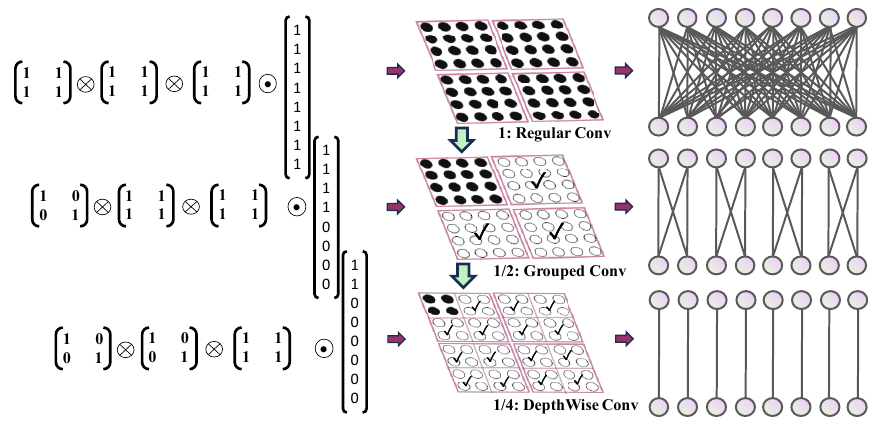}
    \caption{{\small The generation process of GPConv from the core part of DGM, where operator 1 denotes elementwise product,operate 2 denotes a Kronecker product from left to right.}}
    \label{fig:GPConv}
\end{figure}

In essence, the interplay between DCL and DGM constitutes a hierarchical decoding process, in which the model sequentially reconstructs and interprets the multi-scale spatial structure and temporal progression information embedded by ARP in the 2D projection. Through this process, diagnostically critical features are progressively recovered and emphasized during training, effectively overcoming the entanglement and information loss often encountered in direct 2D modeling.

It is crucial to note that the Dynamic Group Mechanism (DGM) operationalises the adaptive nature of DCL. At each curriculum stage, the DGM quantitatively recalibrates channel- and spatial-level attention based on the evolving complexity of features. This ensures that the model's focus remains dynamically aligned with the most salient anatomical or pathological cues throughout the learning process. Such dynamic recalibration is pivotal in achieving the full potential of staged curriculum learning in heterogeneous brain imaging data. The stepwise implementation of DGM is detailed in Algorithm~\ref{alg:DGM}, and the difference between standard and grouped convolution is visualized in Figure~\ref{fig:GPConv}. 

\begin{algorithm}[!h]
    \caption{ Dynamic Group Mechanism }
    \label{alg:DGM}
    \renewcommand{\algorithmicrequire}{\textbf{Input:}}
    \renewcommand{\algorithmicensure}{\textbf{Output:}}
    \begin{algorithmic}[1]
        \REQUIRE Feature map $X \in \mathbb{R}^{C \times H \times W}$
        \ENSURE Transformed feature map $Y$
        \STATE \textbf{Parameters:}  
        \STATE $C$: Number of input channels  
        \STATE $r$: Reduction factor  
        \STATE $G$: Number of groups  

        \STATE \textbf{Step 1: Adaptive Channel Weighting}  
        \STATE $C' \gets \max(1, \lfloor C / r \rfloor)$ // Channel reduction  
        \STATE $W \gets \text{GroupedConv}(X, C', G)$ // Feature extraction  
        \STATE $W \gets \text{ReLU}(W)$ // Non-linearity  
        \STATE $W \gets \text{GroupedConv}(W, C, G)$ // Channel expansion  
        \STATE $W \gets \sigma(W)$ // Weight normalization  
        \STATE // Dynamic importance scaling  

        \STATE \textbf{Step 2: Dynamic Feature Recalibration}  
        \STATE $X' \gets X \odot W$ // Adaptive modulation  

        \STATE \textbf{Step 3: Efficient Feature Fusion}  
        \STATE $X'' \gets \text{Pointwise Groups Conv}(X', C, G)$ \STATE // Spatial feature refinement  
        \STATE $Y \gets \text{GroupedPointwiseConv}(X'', C, G)$ \STATE // Final feature aggregation  
        \RETURN $Y$
    \end{algorithmic}
\end{algorithm}


Figure~\ref{fig:super} illustrates the core model design for the decoding phase of DaSE, highlighting the dynamic curriculum learning strategy developed specifically for spatiotemporal information extraction from encoded representations.

\section{Experiments}
In this section, we first describe the datasets, evaluation metrics, and implementation details. Then, we present the comparison with latest state-of-the-art methods, and discuss the ablation results. 

\subsection{Dataset and Performance Metrics}

To comprehensively evaluate the effectiveness and generalizability of the proposed DCL-SE framework, we conducted experiments on a range of public datasets covering classification, segmentation, and regression tasks in brain imaging (A detailed presentation of the dataset can be found in Supplementary Figures 1-7). While classification serves as the main evaluation scenario, segmentation and brain age prediction tasks are included to demonstrate the extensibility of DCL-SE to broader multi-task applications in neuroimaging.

For classification, four datasets were used. The first is a 2D MRI-based Alzheimer's disease dataset from Kaggle, comprising a total of 40,584 images (33,820 for training/validation, 6,764 for testing). The second and third datasets are multimodal brain tumor classification datasets from Kaggle: BT-MRI (3,278 training/validation images, 655 test images) and BT-CT (4,034 training/validation images, 805 test images). The fourth dataset is a 3D NIfTI-format Alzheimer's disease dataset from the ADNI (Alzheimer’s Disease Neuroimaging Initiative) database, used for binary classification of 3D volumes.

To further demonstrate the multi-task capability of DCL-SE, we evaluate vessel segmentation and brain age prediction. For segmentation, the CAS2023 dataset from the Cerebral Artery Segmentation Challenge 2023 is used. This dataset contains 100 annotated 3D TOF-MRA (Time-of-Flight Magnetic Resonance Angiography) scans, with voxel dimensions ranging from (208, 320, 96) to (784, 784, 255). A total of 14,349 slices are available, of which 1,400 containing vessel regions were selected for analysis. In segmentation labels, 0 denotes background and 1 denotes cerebral arteries. 

For brain age prediction, we utilize the SPR Head CT Age Prediction Challenge dataset, organized by the Radiology and Diagnostic Imaging Society of São Paulo (SPR). This large-scale public dataset consists of 525,337 head CT scans in DICOM format (total size 276.93 GB), with accompanying patient age information provided in CSV files. The challenge is designed to develop models capable of accurately predicting the patient's age group from head CT images.

Classification performance is measured using accuracy (Acc), area under the receiver operating characteristic curve (AUC), F1-score (F1), precision (Pre), recall (Rec), average precision (AP), number of model parameters (\#Params), and floating point operations per second (GFLOPS). For segmentation, the Dice Similarity Coefficient (DSC, also referred to as DICE) is adopted as the main evaluation metric. For brain age prediction, mean absolute error (MAE) is used.

The evaluation metrics are defined as follows:

\begin{itemize}
    \item {Accuracy (Acc):}
    \begin{equation}
        \mathrm{Acc} = \frac{TP + TN}{TP + TN + FP + FN}
    \end{equation}
    where $TP$, $TN$, $FP$, and $FN$ denote true positives, true negatives, false positives, and false negatives, respectively.
    \item {Precision (Pre):}
    \begin{equation}
        \mathrm{Pre} = \frac{TP}{TP + FP}
    \end{equation}
    \item {Recall (Rec):}
    \begin{equation}
        \mathrm{Rec} = \frac{TP}{TP + FN}
    \end{equation}
    \item {F1-score (F1):}
    \begin{equation}
        \mathrm{F1} = \frac{2 \cdot \mathrm{Pre} \cdot \mathrm{Rec}}{\mathrm{Pre} + \mathrm{Rec}}
    \end{equation}
    \item {Dice Similarity Coefficient (DSC):}
    \begin{equation}
        \mathrm{DSC} = \frac{2 |P \cap G|}{|P| + |G|}
    \end{equation}
    where $P$ and $G$ are the sets of predicted and ground truth positive voxels, respectively.
    \item {Mean Absolute Error (MAE):}
    \begin{equation}
        \mathrm{MAE} = \frac{1}{N} \sum_{i=1}^{N} |y_i - \hat{y}_i|
    \end{equation}
    where $y_i$ and $\hat{y}_i$ denote the true and predicted ages, and $N$ is the number of samples.
\end{itemize}

These diverse tasks collectively highlight not only the accuracy and efficiency of DCL-SE in standard classification, but also its ability to generalize and adapt to complex multi-task scenarios in neuroimaging analysis.

\subsection{Implementation Details}

\begin{table*}[]
    \centering
    \setlength{\abovecaptionskip}{0cm} 
    \caption{Performance Comparison of Different Pooling Methods on AD 3D Dataset Using DCL-SE Model}
    \setlength{\tabcolsep}{4.62mm}{
    \begin{tabularx}{14cm}{ccccccc}
        \hline
        \textbf{Method} &
        \multicolumn{1}{c}{\textbf{Acc}} &
        \multicolumn{1}{c}{\textbf{Auc}} &
        \multicolumn{1}{c}{\textbf{F1}} &
        \multicolumn{1}{c}{\textbf{Pre}} &
        \multicolumn{1}{c}{\textbf{Recall}} &
        \multicolumn{1}{c}{\textbf{Ap}} \\
        \hline
        Max Pooling           & 88.5\%  & 92.3\%   & 89.2\% & 85.2\%   & 93.5\% & 83.0\%   \\ 
        Mean Pooling            & 90.1\% & 89.2\%   & 90.0\% & 93.1\%   & 87.1\% & 87.6\%   \\
        Global Average Pooling           & 85.2\%    & 89.5\%   & 84.7\%    & 89.2\%   & 80.6\%    & 81.8\%  \\
        Stochastic Pooling               & 88.5\% & 88.3\%   & 88.1\% & 92.8\%   & 83.8\% & 86.0\%   \\
        Spatial Pyramid Pooling                  & 88.5\%    & 91.9\%   & 88.5\%    & 90.0\%   & 87.1\%    & 84.9\%   \\
        \rowcolor{gray!20} 
        Approximate Rank Pooling          & \textbf{96.4\%}    & \textbf{97.1\%}   & \textbf{96.6\%}    & \textbf{96.6\%}   & \textbf{96.6\%}    & \textbf{95.2\%}   \\
        \hline
        \label{Table:Pool}
    \end{tabularx} }
    \vspace{-2.0em}
\end{table*}

\begin{table*}[htbp] 
\setlength{\abovecaptionskip}{0cm} 
\centering
\caption{The Performance of Each Model of Ablation Experiment} 
\setlength{\tabcolsep}{1.8mm}
\begin{tabular}{ccccccccccccccc}
\hline
\multicolumn{4}{c}{\textbf{Stage}} & \multicolumn{1}{l}{} & \multicolumn{3}{c}{\textbf{Paradigm}} & \multicolumn{1}{l}{} & \multicolumn{6}{c}{\textbf{Evaluation}} \\ \cline{1-4} \cline{6-8} \cline{10-15} 
\textbf{\raisebox{-0.2ex}{S1}}  &
\textbf{\raisebox{-0.2ex}{S4}} &
\textbf{\raisebox{-0.2ex}{S6}} &
\textbf{\raisebox{-0.2ex}{C1}} &
  \multicolumn{1}{l}{\textbf{}} &
  \textbf{\raisebox{-0.2ex}{Dynamic}} &
  \textbf{\raisebox{-0.2ex}{CL}} &
  \multicolumn{1}{l}{\textbf{\raisebox{-0.2ex}{DCL}}} &
  \multicolumn{1}{l}{\textbf{}} &
  \multicolumn{1}{l}{\textbf{\raisebox{-0.2ex}{F1}}} &
  \multicolumn{1}{l}{\textbf{\raisebox{-0.2ex}{Acc}}} &
  \multicolumn{1}{l}{\textbf{\raisebox{-0.2ex}{Auc}} }&
  \multicolumn{1}{l}{\textbf{\raisebox{-0.2ex}{Ap}} }&
  \multicolumn{1}{l}{\textbf{\raisebox{-0.2ex}{\#params}}} &
  \multicolumn{1}{l}{\textbf{\raisebox{-0.2ex}{GFLOPS}} }\\ \hline
            $\checkmark$ & $\checkmark$ & $\checkmark$ & $\checkmark$ &  &  $\times$ & $\checkmark$ & $\times$ &   &   93.6\% & 93.0\% & 96.4\% & 89.4\% & 7.43M  & 1.09G  \\
            $\times$ & $\checkmark$ & $\checkmark$ & $\checkmark$ &   &    $\checkmark$ & $\times$ & $\times$ &    &  90.9\% & 89.4\% & 91.1\% & 83.3\% & 8.060M & 1.126G \\
            $\checkmark$ & $\times$ & $\checkmark$ & $\checkmark$ &   &    $\checkmark$ & $\times$ & $\times$ &    &  90.0\% & 89.4\% & 93.3\% & 86.2\% & 8.059M & 1.127G \\
            $\checkmark$ & $\checkmark$ & $\times$ & $\checkmark$ &   &   $\checkmark$ & $\times$ & $\times$ &    &  88.1\% & 87.7\% & 90.0\% & 84.7\% & 8.051M & 1.127G \\
            $\checkmark$ & $\checkmark$ & $\checkmark$ & $\times$ &   &    $\checkmark$ & $\times$ & $\times$ &    &  89.6\% & 89.4\% & 93.2\% & 87.4\% & 7.446M & 1.097G \\
            
            \rowcolor{gray!20}
            $\checkmark$ & $\checkmark$ & $\checkmark$ & $\checkmark$ &  &
            $\checkmark$ & $\checkmark$ & $\checkmark$ &  &
            \textbf{96.6\%} & \textbf{96.4\%} & \textbf{97.1\%} & \textbf{95.2\%} & 8.060M & 1.127G \\
\hline
\end{tabular}
\label{Table:Ablation}
\vspace{-1.0em}
\end{table*}

In our experiments, we primarily used GeForce RTX 3090 and 4090 GPUs equipped with 24GB of memory for brain image classification tasks. As a multi-task extension, We also used NVIDIA Tesla A100 and P100 GPUs for training brain tumor segmentation and comparative experiments on brain age prediction.

For classification tasks, we use the AdamW optimiser with weight decay (0.01) is used in conjunction with a cosine annealing learning rate strategy (initially set to 0.001 and reset every 50 epochs) in order to avoid local optima. 2D data is trained for 300 epochs (batch size 64), while 3D data uses 3-fold cross-validation (200 epochs per fold).

For segmentation tasks, we made targeted adjustments to the classification task training framework to accommodate the characteristics of pixel-level recognition: The AdamW optimiser (weight decay 0.0001, initial learning rate 0.0001) is used in conjunction with a cosine annealing restart (cycle of 10 epochs) and a ReduceLROnPlateau strategy (50\% decay after 5 epochs without improvement). The loss function is a weighted combination of 0.5 × Lovász-Softmax loss (category balance), 0.3 × Lovász-Hinge loss (region overlap) and 0.2 × boundary loss (edge accuracy).

The prediction tasks, we use an end-to-end training approach with the AdamW optimiser (learning rate 5e-5, weight decay 1e-5) is optimised using L1 loss (mean absolute error (MAE)) and a OneCycleLR scheduler with 5 epoch warm-up, followed by cosine annealing to 1e-6 to balance convergence speed and parameter tuning.

\subsection{Ablation Studies}
The DCL-SE framework is built on the Data-based Spatiotemporal Encoding (DaSE) paradigm, which integrates ARP-based encoding and dynamic curriculum-driven decoding. To clarify the contribution of each component within this tightly-coupled framework, we conduct ablation experiments on the 3D Alzheimer's disease dataset, focusing on three key aspects: spatiotemporal encoding (ARP), staged curriculum decoding (DCL-SE), and dynamic feature recalibration (DGM).

\begin{table*}[] 
    \centering 
    \setlength{\abovecaptionskip}{0cm}
    \caption{Model Performance Compared on Alzheimer’s disease(AD) and Brain Tumor 2D Dataset}
    \setlength{\tabcolsep}{3.3mm}{
    \begin{tabularx}{14cm}{cccccc}
        \hline
        \textbf{Method} & \textbf{AD-MRI/Acc} & \textbf{BT-MRI/Acc} & \textbf{BT-CT/Acc} & \textbf{\#params} & \textbf{GFLOPS} \\
        \hline
        SwinT          & 98.00\% & 97.23\% & 98.21\% & 87.70M  & 15.16G \\
        ConvNeXt       & 99.36\% & 98.78\% & 98.14\% & 88.54M  & 15.35G \\
        NiFuse(2D-SOTA)         & 99.88\% & 97.02\% & \textbf{98.78\%} & 13.30M  & 20.23G \\
        Avg.+VGG11+Att\cite{adeccv} & 99.80\% & \textbf{99.85\%} & 92.68\% & 121.87M & 30.13G \\
        DAMNet(2D,3D-SOTA)\cite{ad2025}         & 99.90\% & 99.69\% & 98.01\% & 7.4M    & 1.1G   \\
        \rowcolor{gray!20} 
        DCL-SE       & \textbf{99.94\%} & \textbf{99.85\%} & 97.89\% & 8.0M    & 1.1G   \\
        \hline
    \label{Table:2D}
    \end{tabularx} }
    \vspace{-2.0em}
\end{table*}

\begin{table*}[] 
    \centering
    \setlength{\abovecaptionskip}{0cm} 
    \caption{Model Performance Compared on Alzheimer’s disease(AD) 3D Dataset. (\#BE:Best Epoch)}
    \setlength{\tabcolsep}{1.82mm}
    \begin{tabularx}{14cm}{ccccccccc}
        \hline
        \textbf{3D-NII/Method} &
        \multicolumn{1}{c}{\textbf{Acc}} &
        \multicolumn{1}{c}{\textbf{Auc}} &
        \multicolumn{1}{c}{\textbf{F1}} &
        \multicolumn{1}{c}{\textbf{Pre}} &
        \multicolumn{1}{c}{\textbf{Recall}} &
        \multicolumn{1}{c}{\textbf{Ap}} &
        \multicolumn{1}{c}{\textbf{\#params}} &
        \multicolumn{1}{c}{\textbf{GFLOPS}} \\
        \hline
        AlexNet\cite{adeccv}              & 87\%  & 90\%   & 86\% & 89\%   & 83\% & 82\%  & 61.10M & 0.71G \\ 
        MobileNetV2\cite{adeccv}            & 88\%    & 89\%   & 87\%    & 89\%   & 85\%    & 83\%   & 3.50M & 0.33G\\
        ConvNeXt               & 57.8\% & 62.3\%   & 64.7\% & 57.8\%   & 73.3\% & 56.4\%  & 88.54M & 15.35G \\
        NiFuse(2D-SOTA)                  & 73.6\%    & 72.5\%   & 73.6\%    & 77.7\%   & 70.0\%    & 70.2\%  & 13.30M & 20.23G \\
        Avg.+VGG11+Att\cite{adeccv}            & 88.0\%    & 89.0\%   & 88.0\%    & 85.0\%   & 91.0\%    & 82.0\%  & 121.87M & 30.13G \\
        DAMNet(2D,3D-SOTA)\cite{ad2025}                 & 93.0\%  & 96.4\% & 93.6\%  & 90.6\% & 96.7\%  & 89.4\% & 7.4M & 1.1G\\
        \hline
        \rowcolor{gray!20}
        Fold1-(\#BE=32)  & 100.0\% & 100.0\% & 100.0\% & 100.0\% & 100.0\% & 100.0\% & 8.0M & 1.1G \\
        \rowcolor{gray!20}
        Fold2-(\#BE=47)  & 94.1\% & 95.4\% & 95.2\% & 100.0\%  & 90.9\% & 96.7\% & 8.0M & 1.1G \\
        \rowcolor{gray!20}
        Fold3-(\#BE=51)  & 95.2\% & 98.1\% & 95.6\% & 91.6\% & 100.0\% & 91.6\% & 8.0M & 1.1G \\
        \rowcolor{gray!20}
        DCL-SE           & \textbf{96.4\%} & \textbf{97.1\%} & \textbf{96.6\%} & \textbf{96.6\%} & \textbf{96.6\%} & \textbf{95.2\%} & 8.0M & 1.1G \\
        \hline
        \label{Table:3D}
    \end{tabularx} 
    \vspace{-2.0em}
\end{table*}

We first evaluate the role of Approximate Rank Pooling (ARP) in DaSE’s encoding stage. Unlike order-invariant pooling (e.g., mean or max), ARP preserves anatomical progression and global structural cues, which are essential for subject-level diagnosis. As shown in Table~\ref{Table:Pool} and Supplementary Tables 1-5, replacing ARP with standard pooling significantly reduces accuracy and F1-score (e.g., F1 drops from 96.6\% to 89.2\% for max pooling, and to 84.7\% for global average pooling). The AUC with ARP reaches 97.1\%, substantially outperforming alternatives. This confirms that order-sensitive aggregation is indispensable for encoding clinically relevant spatiotemporal information.

For the decoding stage, we analyze the impact of the dynamic curriculum mechanism. Removing the curriculum strategy causes the F1-score to drop from 96.6\% to 88.1\%, and AUC from 97.1\% to 90.0\% (Table~\ref{Table:Ablation}), indicating that progressive, staged learning is critical for disentangling complex features and aligning model focus with the hierarchical nature of neuroanatomical progression.

The Dynamic Group Mechanism (DGM) further enhances decoding by adaptively recalibrating channel and spatial attention at each curriculum stage. Removing DGM from any stage results in consistent drops in all metrics, showing that static attention is inadequate for capturing diagnostically subtle features. Partial removal of curriculum stages (e.g., omitting S4 or S6) also degrades performance (F1 and AUC decrease by 6–8\%), indicating that multi-scale feature fusion is essential for robust discrimination.

\subsection{Comparative Studies}
To rigorously assess the effectiveness and generalizability of DCL-SE, we benchmarked the framework on four representative neuroimaging datasets, covering both 2D (AD-MRI, BT-MRI, BT-CT) and 3D (AD-NII) classification tasks. For fair comparison, state-of-the-art 2D and 3D methods—including SwinT\cite{swin}, Avg.+VGG11+Att\cite{adeccv}, DAMNet\cite{ad2025}, NiFuse\cite{hifuse} and ConvNeXt\cite{convnext} and classic CNNs—were used as baselines.

Experimental results (Table~\ref{Table:2D} and Table~\ref{Table:3D}) demonstrate that DCL-SE consistently matches or outperforms all baselines across both 2D and 3D settings. Notably, on the AD-NII volumetric dataset, DCL-SE achieves 96.4\% accuracy, 97.1\% AUC, and a 96.6\% F1-score, surpassing advanced 2D-centric and hybrid models, which suffer significant performance drops in 3D contexts. This advantage is directly attributed to the DaSE paradigm: Stage 1 (ARP encoding) preserves spatial progression and anatomical context, while Stage 2 (curriculum-driven decoding with DGM) incrementally disentangles and emphasizes diagnostically relevant features. The synergy between order-sensitive encoding and dynamic, complexity-aware decoding is essential for robust volumetric brain analysis.

\begin{table}[h]
\setlength{\abovecaptionskip}{0cm} 
\centering
\caption{Comparison and analysis with SOTA works on AD using the DaSE processing method.} 
\begin{tabular}{lllllll}
\hline
\raisebox{0pt}{Models}  &  & \multicolumn{2}{c}{DaSE Encapsulation Idea} &  & \multicolumn{2}{c}{Evaluation} \\ \cline{3-4} \cline{6-7} 
 &
   &
  \begin{tabular}[c]{@{}c@{}}\#DC\\ Data attributes\end{tabular} &
  \begin{tabular}[c]{@{}c@{}}\#CS\\ Data Training\end{tabular} &
   &
  \begin{tabular}[c]{@{}c@{}}F1\\ (\%)↑\end{tabular} &
  \begin{tabular}[c]{@{}c@{}}Auc\\   (\%)↑\end{tabular} \\ \cline{1-1} \cline{3-4} \cline{6-7} 
ResNet18                &  & \makecell[r]{$\times$}                    & \makecell[r]{$\times$}                   &  & 84.0           & 84.0          \\
DAMNet                  &  & \makecell[r]{$\checkmark$}                    & \makecell[r]{$\times$}                    &  & 93.6           & 96.4          \\
\cellcolor{gray!20}DCL-SE          &  & \makecell[r]{$\checkmark$}                    & \makecell[r]{$\checkmark$}                    &  & \cellcolor{gray!20}96.6           & \cellcolor{gray!20}97.1          \\ \hline
\label{Table:compareDaSE}
\end{tabular}
\vspace{-1.0em}
\end{table}

The comparative effect of DaSE is further quantified in Table~\ref{Table:compareDaSE}: only DCL-SE fully implements both ARP encoding and curriculum-based decoding, leading to the best F1 and AUC scores (96.6\%, 97.1\%) compared to DAMNet (which lacks dynamic curriculum) and ResNet18 (which lacks both). This validates that both structural encoding and staged, adaptive feature refinement are critical to effective multimodal medical image modeling.

\begin{figure}[htbp]
  \centering
  \subfloat[]{  
      \includegraphics[width=0.45\textwidth]{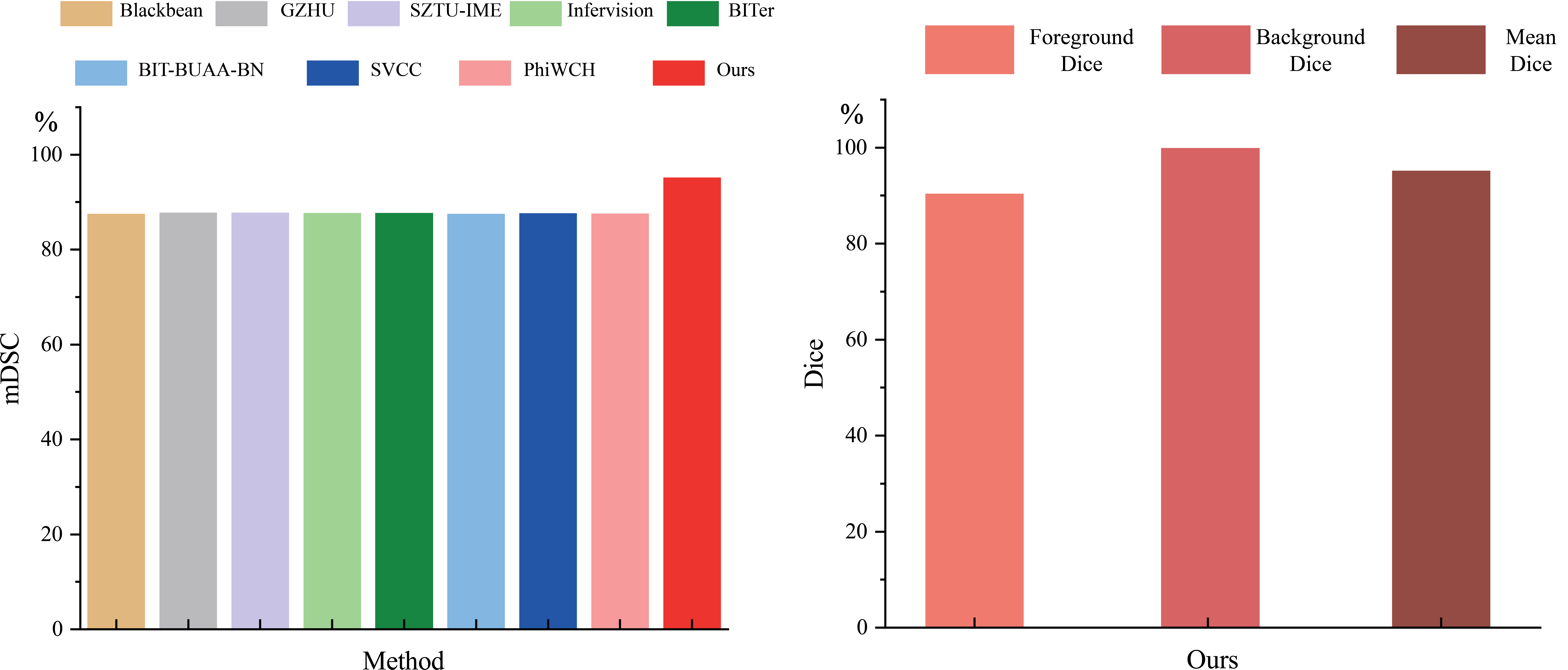}
      \label{fig:seg1}
  } \\   
  \subfloat[]{  
      \includegraphics[width=0.45\textwidth]{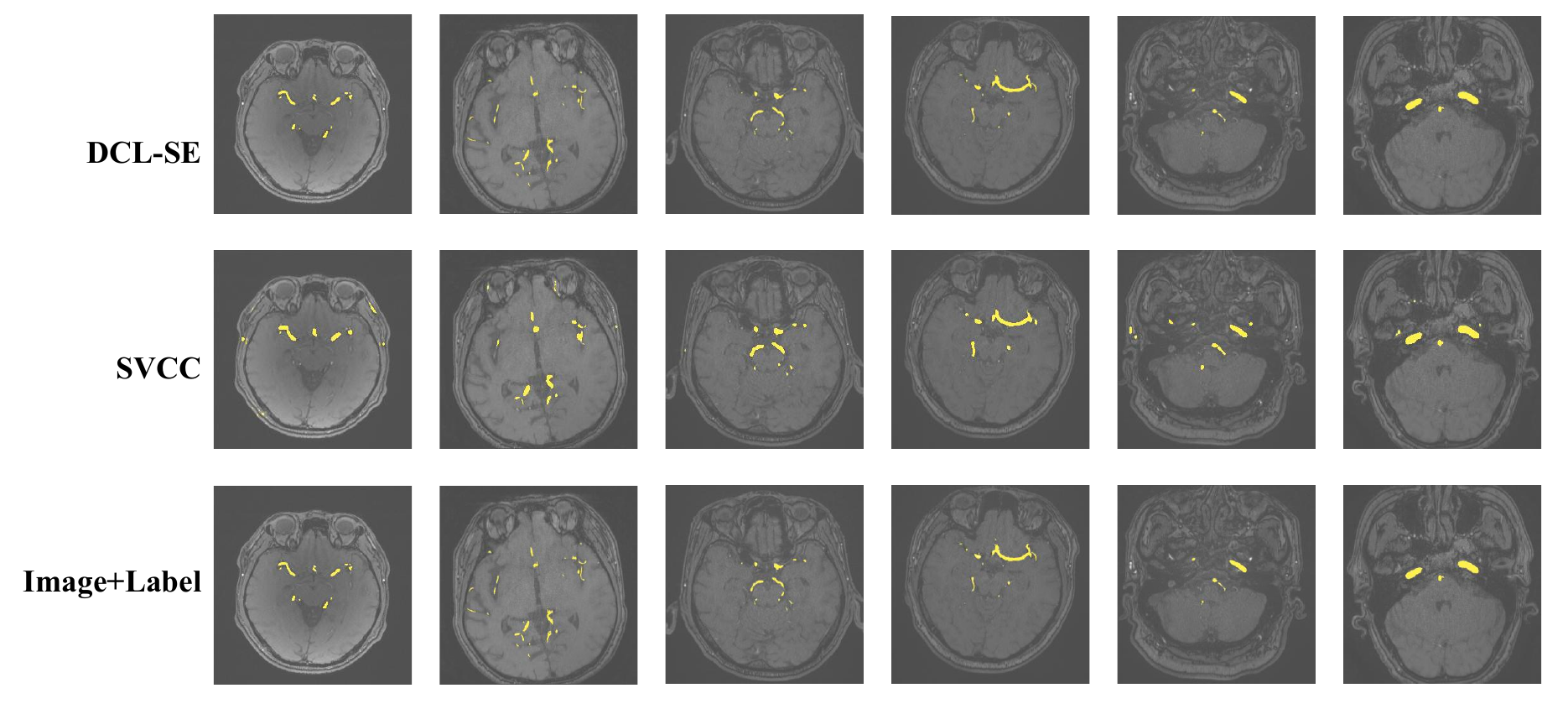}
      \label{fig:seg2}
  }
  \caption{(a) Vascular segmentation on the CAS2023 3D-MRA (.nii.gz) dataset: The left panel presents mean Dice similarity coefficients (mDSC) for eight methods—Blackbean, GZHU, SZTU-IME, Infervision, BITer, BIT-BUAA-BN, SVCC, and PhiWCH—compared with DCL-SE. The right panel reports DCL-SE’s Dice scores for vessel tissue (DSC 1), background (DSC 0), and their mean. All experiments used .nii.gz volumes from the CodaLab Competition (https://codalab.lisn.upsaclay.fr/competitions/9804); full results are in Supplementary Table~6. (b) Qualitative comparison on CAS2023 3D-MRA: columns show different axial slices, with rows for DCL-SE (top), SVCC (middle), and ground-truth annotations (bottom, yellow overlay).}
  \label{fig:seg}
\vspace{-1.0em}
\end{figure}

\begin{figure}[htbp]
\setlength{\abovecaptionskip}{0cm} 
    \centering
    \includegraphics[width=0.45\textwidth]{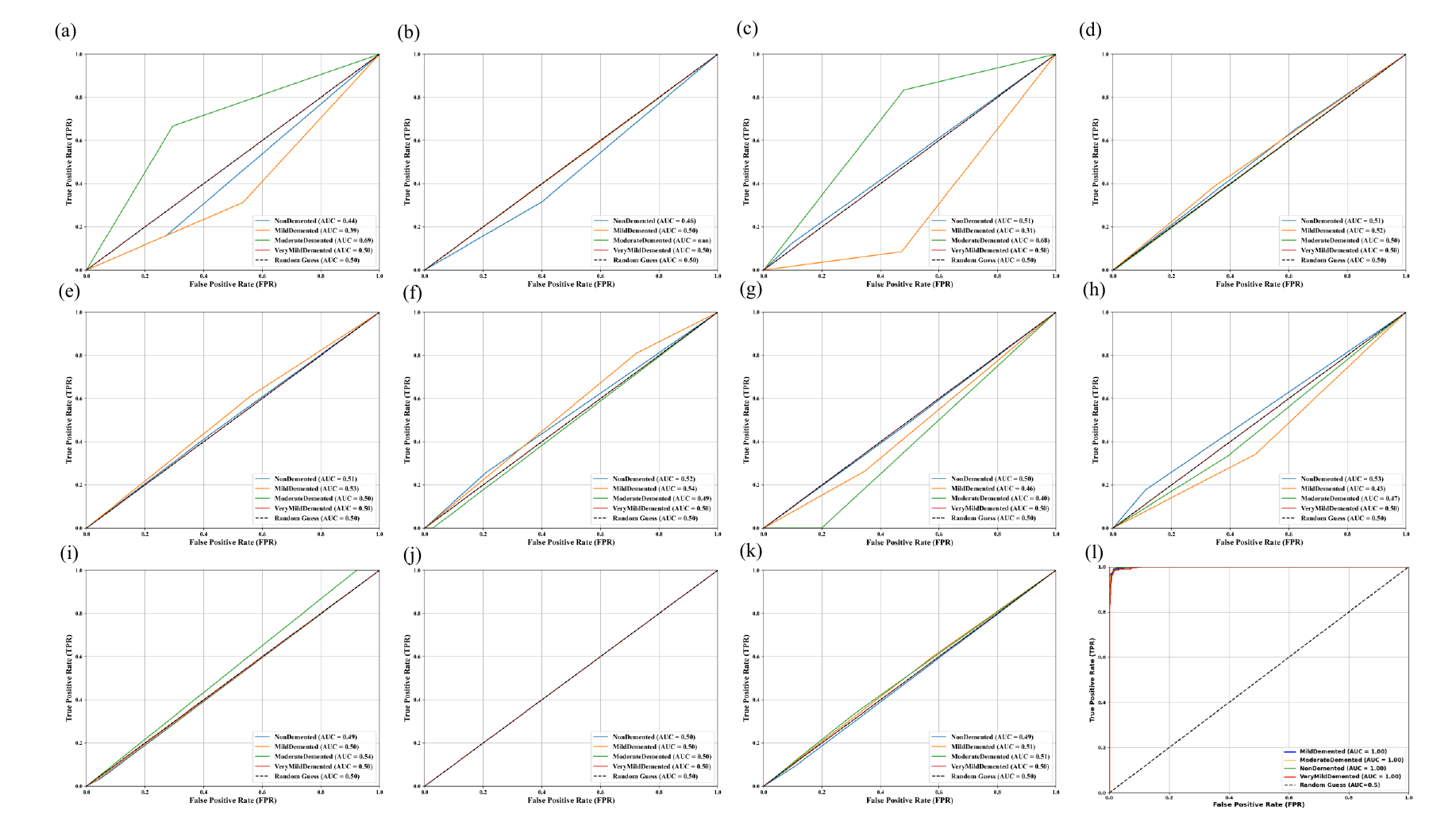}
  \caption{Receiver‐operating characteristic (ROC) curves for eleven off‐the‐shelf, non‐fine-tuned large language models—ERNIE-Speed-128K (a), claude-3-5-haiku (b), claude-3-haiku-20240307 (c), flux (d), gemini-2.0-flash-thinking-exp (e), glm-4-flash (f), gpt-4o (g), gpt4 (h), hunyuan-lite (i), moonshot-v1-8k-vision-preview (j) and pixtral-12b-2409 (k)—alongside our proposed compact model DCL-SE (l). The dashed diagonal line represents random‐guess performance (AUC = 0.5).}
  \label{fig:auc}  
\vspace{-1.0em}
\end{figure}

\subsection{Expansion Studies}
To evaluate the multi-task generalizability of DCL-SE, we conducted experiments on cerebral artery segmentation and brain age prediction, extending beyond standard classification tasks.

For brain vessel segmentation, DCL-SE was benchmarked on the CAS2023 dataset, achieving the highest mean Dice similarity coefficient (mDSC) among all competing methods (Fig.\ref{fig:seg1}), and demonstrating more anatomically precise boundaries than strong baselines (Fig.\ref{fig:seg2}).  More details can be found in Supplementary Figures 8-12.  For brain age prediction, DCL-SE outperformed six widely-used open-source models on the SPR Head CT Age Prediction Challenge dataset, achieving the lowest mean absolute error (MAE) despite large-scale data heterogeneity. Notably, the MAE of DCL-SE was 9.47 lower than that of FiA-Net, which previously exhibited strong performance on this task (see Supplementary Table 7 for full results).

To further contextualize the strengths and limitations of specialized models like DCL-SE in the era of foundation models, we additionally conducted a systematic comparison between 11 representative large-scale pre-trained models and lightweight task-specific networks on the 2D Alzheimer's disease classification benchmark. Importantly, all large models were evaluated in a zero-shot manner, without task-specific fine-tuning, reflecting the realities of resource-limited clinical environments. As illustrated in Fig.~\ref{fig:auc} and detailed in Supplementary Tables 8–10, small models such as DCL-SE not only maintained consistently high accuracy and AUC (with Val\_Acc and Test\_Acc reaching up to 97.81\% and 97.71\%, and even 100\% AUC in some cases), but also outperformed large models, whose predictions exhibited greater variability and, in some instances, approached random chance. This pronounced performance contrast on a relatively simple four-class classification task underscores the continuing necessity of compact, specialized architectures in foundational clinical applications, even as large models proliferate. These findings highlight the potential for complementarity between large-scale and small-scale models in practical medical AI systems, suggesting that lightweight models remain indispensable for efficient and robust deployment in real-world clinical scenarios.

\subsection{Interpretability}

\begin{figure}[htbp]
  \centering
  \subfloat[]{  
      \includegraphics[width=0.45\textwidth]{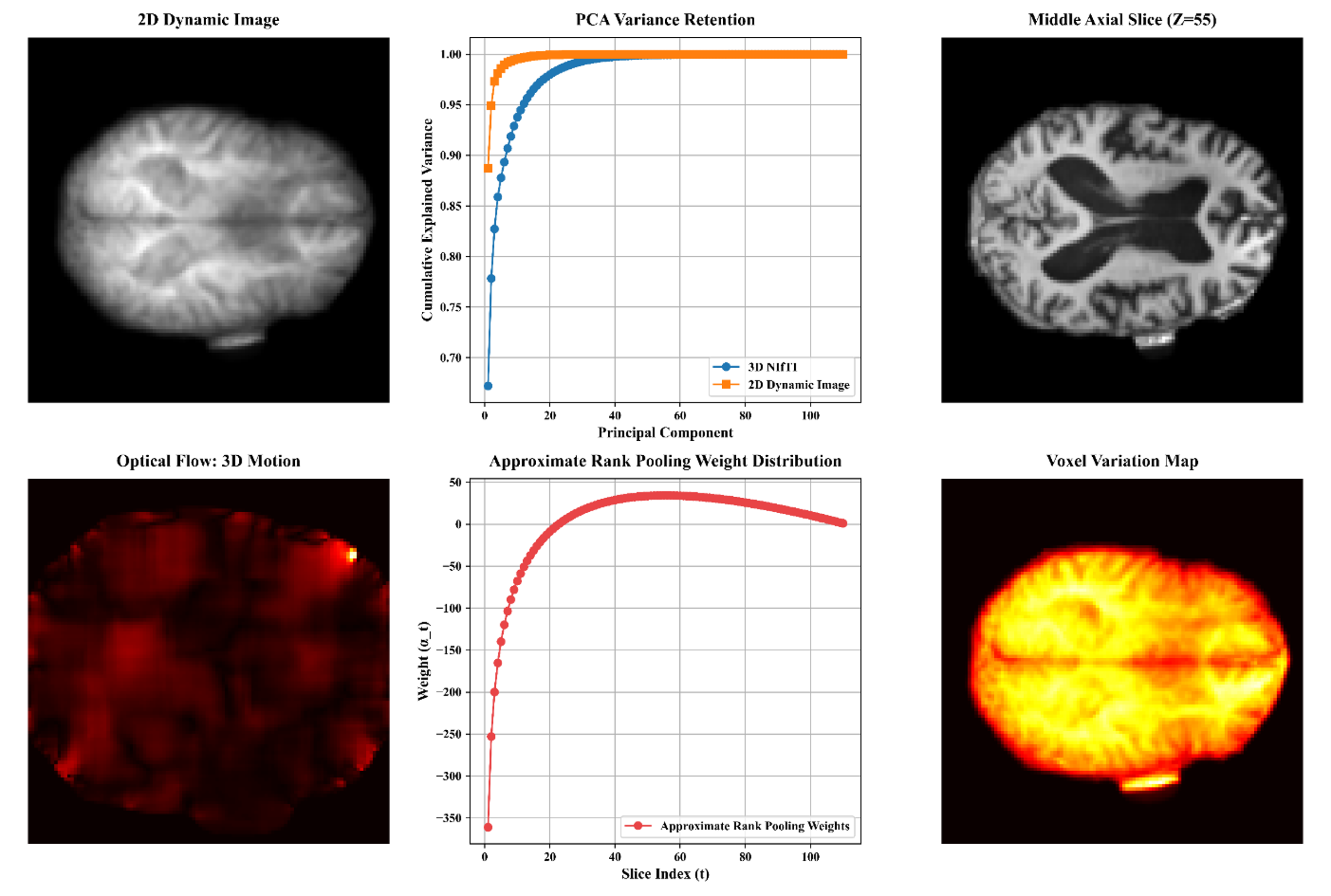}
      \label{fig:arpa}
  } \\   
  \subfloat[]{  
      \includegraphics[width=0.45\textwidth]{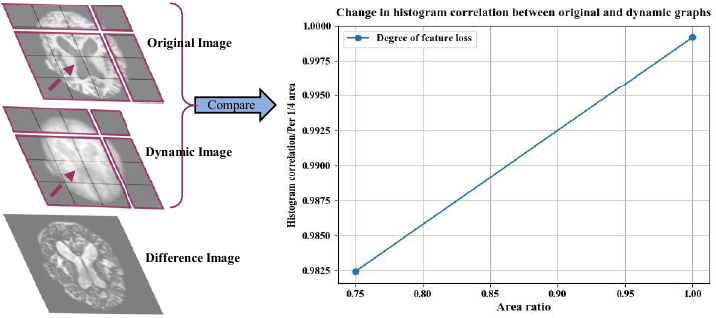}
      \label{fig:arpb}
  }
  \caption{Interpretability of feature loss in image format conversion. (a) Dynamic information analysis and visualization of 2D-3D conversion; (b) Histogram correlation-based feature loss degree for data format conversion.}
  \label{fig:arp}
\vspace{-1.0em}
\end{figure}

A major challenge in deep learning-based neuroimaging is model interpretability, particularly for complex transformations such as 3D-to-2D compression and multi-stage feature decoding. In DCL-SE, interpretability arises directly from the explicit architectural design of DaSE, which consists of two tightly coupled stages: (1) ARP-based encoding, and (2) dynamic curriculum decoding via DGM. We conduct both quantitative and visual analyses to clarify how each stage preserves and structures diagnostically relevant information.

The first stage, ARP encoding, projects 3D brain MRI volumes into 2D dynamic images while retaining essential spatial and progression cues. As demonstrated in Fig.\ref{fig:arpa}, the ARP-generated representations maintain both global and local structural features, effectively filtering out noise. Principal component analysis confirms that the major axes of anatomical variation are preserved in the ARP outputs, with minimal loss limited to higher-order components. Histogram correlation analysis (Fig.\ref{fig:arpb}) further quantifies this, indicating high fidelity between the original volumes and their dynamic representations across all modalities. For quantitative assessment, all data are normalized to the $[0, 1]$ range, ensuring consistent cross-modal comparison.

\begin{figure}[htbp]
  \centering
  \subfloat[]{  
      \includegraphics[width=0.45\textwidth]{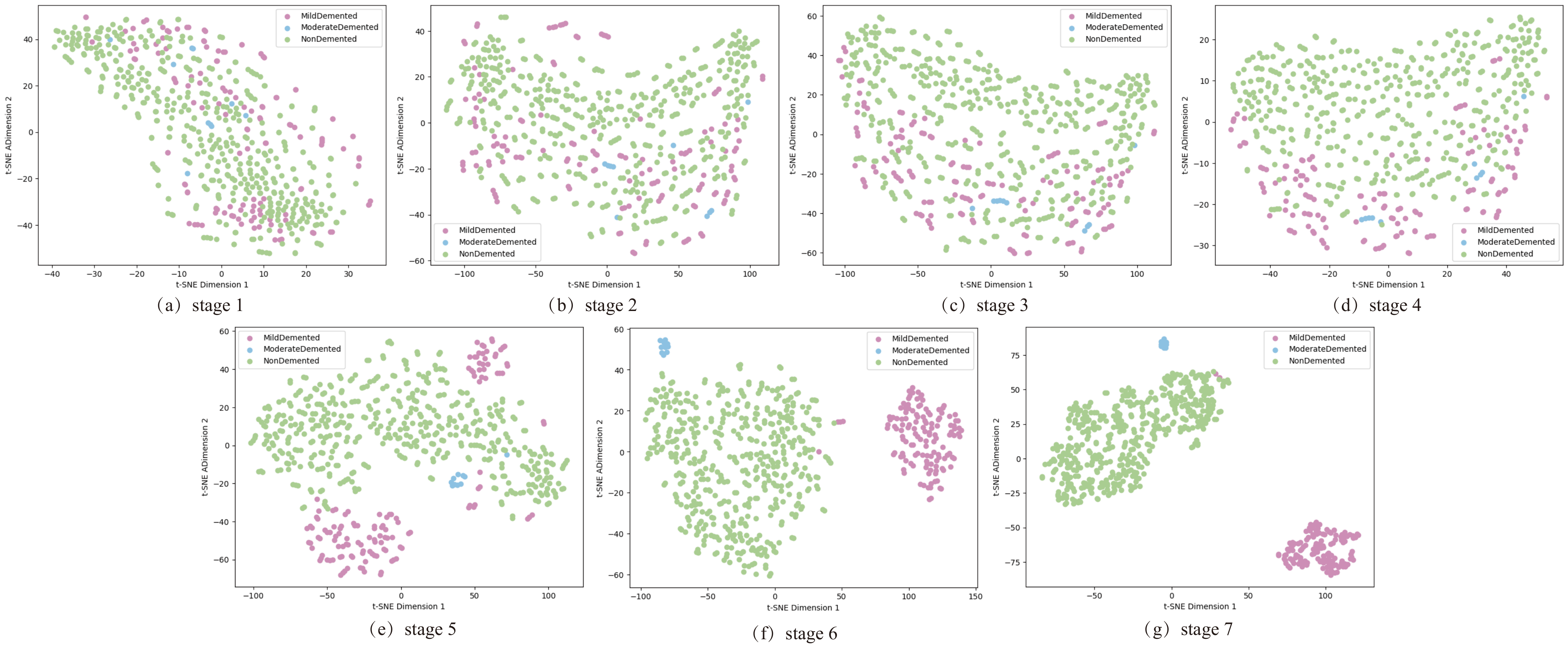}
      \label{fig:tsnea}
  } \\   
  \subfloat[]{  
      \includegraphics[width=0.45\textwidth]{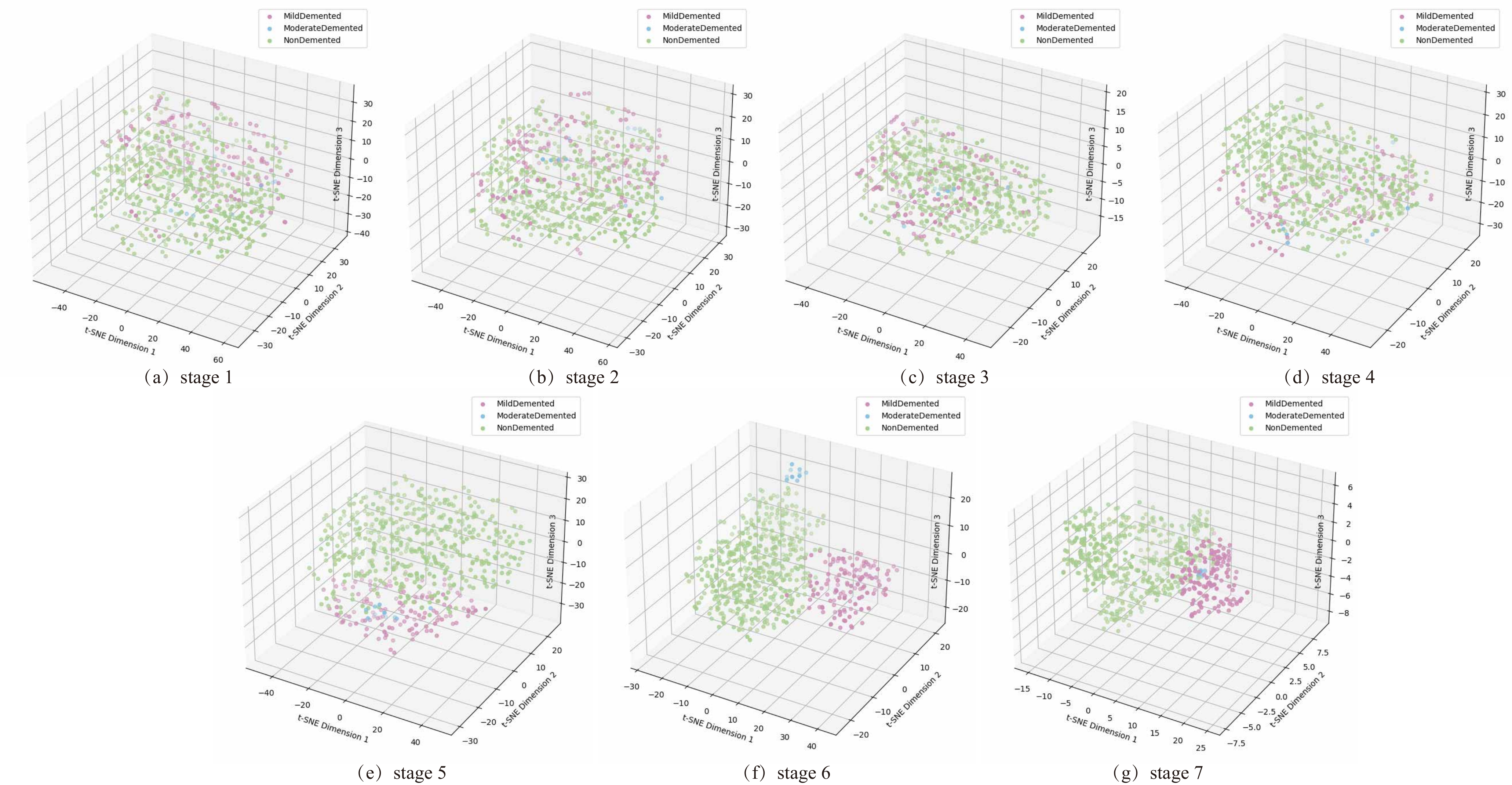}
      \label{fig:tsneb}
  }
  \caption{Visualization of model interpretability across DaSE stages in the classification task: progressive learning from stage 1 to stage 7 transforms Alzheimer's disease classification from dispersed clusters to three clearly separable categories. (a) t-SNE 2D visualization for 2D images; (b) t-SNE 3D visualization for 2D images.}
  \label{fig:vistsne}
\vspace{-1.5em}
\end{figure}

\begin{figure}[htbp]
    \centering
    \includegraphics[width=0.45\textwidth]{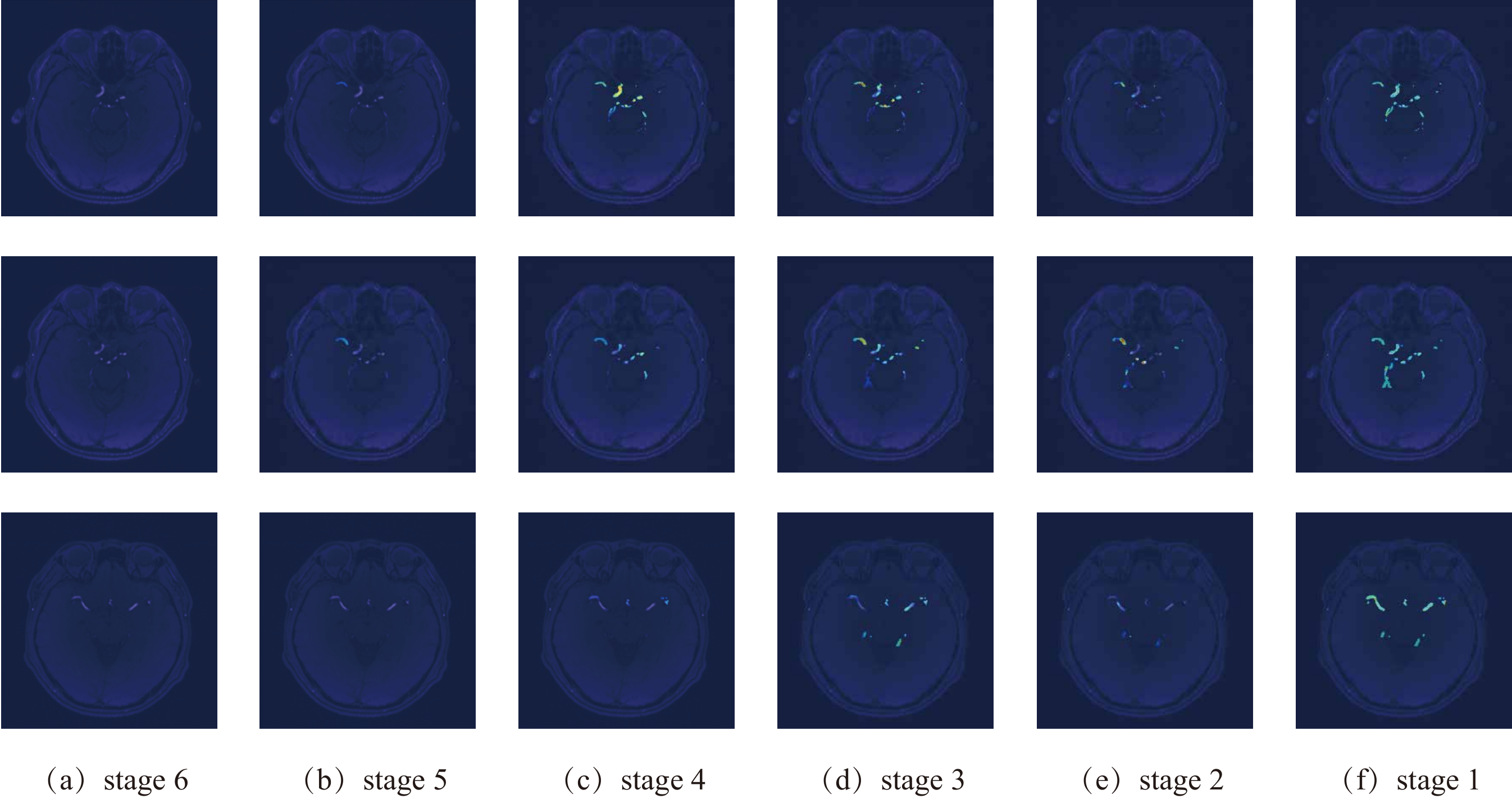}
  \caption{Visualization of the interpretability of each stage of the model based on the DaSE mechanism in segmentation task. Through progressive learning using the DaSE mechanism, from stage 6 to stage 1, the model's segmentation of cerebral arteries evolves from a blurred state to a clearly visible arterial division.}
  \label{fig:visgradcam}  
\vspace{-1.0em}
\end{figure}

The second stage, dynamic curriculum decoding, enables progressive disentanglement of features and hierarchical learning—mirroring the clinical reasoning process from coarse anatomy to fine pathology. t-SNE visualization of feature embeddings at successive curriculum stages (Fig.~\ref{fig:vistsne}) reveals how model representations evolve: early stages show mixed or overlapping clusters, while final stages form clearly separated, clinically meaningful categories. This progressive organization demonstrates that the curriculum-driven decoding (supported by DGM) systematically increases feature separability and transparency throughout the learning process.

In segmentation tasks, GradCAM-based visualizations (Fig.~\ref{fig:visgradcam}) provide additional interpretability: attention maps evolve from diffuse, uncertain focus at early decoding stages to precise anatomical localization at later stages. This stepwise sharpening illustrates how DCL-SE’s staged decoding mechanism achieves both fine discrimination and anatomical consistency without sacrificing context.

These analyses confirm that DCL-SE’s DaSE mechanism—through ARP encoding and curriculum-driven, dynamically recalibrated decoding—achieves interpretable, information-preserving transformation and progressive feature refinement. The method’s internal representations align with clinical reasoning, ensuring both robust performance and practical transparency for subject-level neuroimaging analysis.

\section{Conclusion}
This work presents DCL-SE, a unified framework for subject-level neuroimaging analysis that integrates Data-based Spatiotemporal Encoding (DaSE), dynamic curriculum learning (DCL), and the Dynamic Group Mechanism (DGM). The core paradigm—DaSE—organizes the process into two tightly coupled stages: ARP-based encoding, which preserves essential spatial and progression information from 3D volumetric data, and dynamic curriculum decoding, which enables progressive, adaptive feature refinement via DGM. This design aligns model learning with the hierarchical structure of clinical reasoning, resulting in robust, interpretable extraction of diagnostically salient features.

Comprehensive experiments across multiple neuroimaging benchmarks—including classification, segmentation, and regression tasks—demonstrate that DCL-SE consistently achieves state-of-the-art accuracy, efficiency, and generalizability. Comparative analyses confirm that DCL-SE maintains high performance and interpretability in the face of data heterogeneity, and outperforms both traditional compact models and large-scale foundation models, particularly under real-world clinical constraints.

The staged feature disentanglement and adaptive focus provided by DCL and DGM are central to DCL-SE’s effectiveness, ensuring transparency and reliability in complex clinical scenarios. By bridging the gap between 3D volumetric data and efficient 2D neural architectures, DaSE enables practical cross-modal transfer and progressive training without sacrificing spatial or temporal fidelity.

While this study focuses on neuroimaging, the underlying principles of DCL-SE are broadly applicable. Future work will explore its extension to other imaging domains, integration with external clinical knowledge, and scalable deployment across diverse clinical environments. Collectively, these advances lay the foundation for next-generation, adaptive, and interpretable medical AI systems.

\begin{IEEEbiography}[{\includegraphics[width=1in,height=1.25in,clip,keepaspectratio]{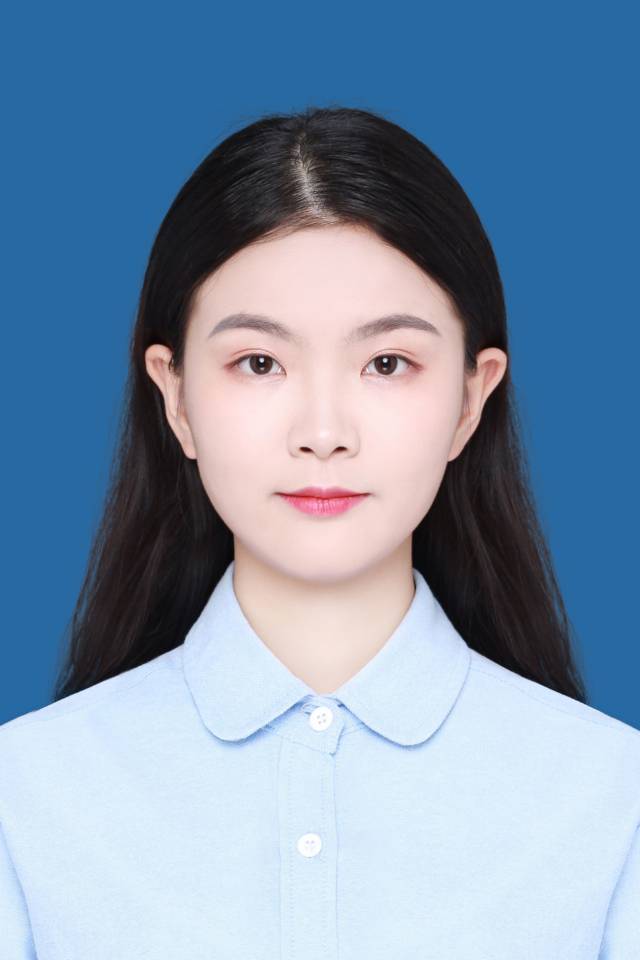}}]{Meihua Zhou} received her Bachelor of Science degree in  Information Management and Information Systems from Wannan Medical university in 2023. From 2024 to 2025, she studied embodied intelligence and multimodal computing in the Department of Computer Science at Tsinghua University, while also acquiring medical knowledge at the Ophthalmology Research Institute of Beijing Tongren Hospital. She is committed to interdisciplinary approaches for solving real-world problems. Currently, she is a graduate student at the University of Chinese Academy of Sciences. Her research interests include, but are not limited to, intelligent medicine, embodied intelligence, and interdisciplinary applications.\end{IEEEbiography}

\begin{IEEEbiography}[{\includegraphics[width=1in,height=1.25in,clip,keepaspectratio]{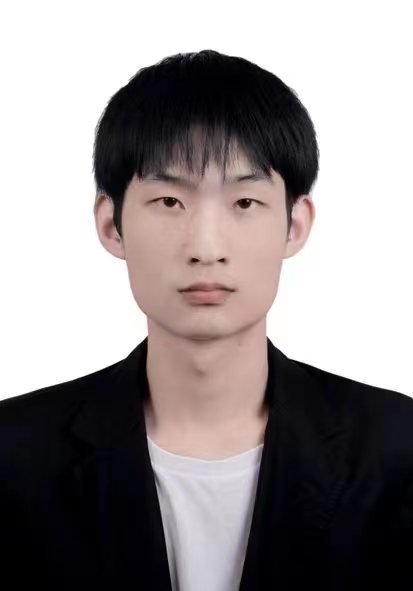}}]{Xinyu Tong} is a graduate student at University of Chinese Academy of Sciences, studying Artificial Intelligence and Biomedical Information Processing. Prior to this he obtained a Bachelor's Degree in Computer Science and his scientific dreams revolved around research in image processing and pattern recognition, using Multimodal-Big data-Efficient algorithms for intelligent medical image analysis and diagnostic interventions. Now his research interests include image processing, pattern recognition, intelligent Ecology and Drug information processing.\end{IEEEbiography}

\begin{IEEEbiography}[{\includegraphics[width=1in,height=1.25in,clip,keepaspectratio]{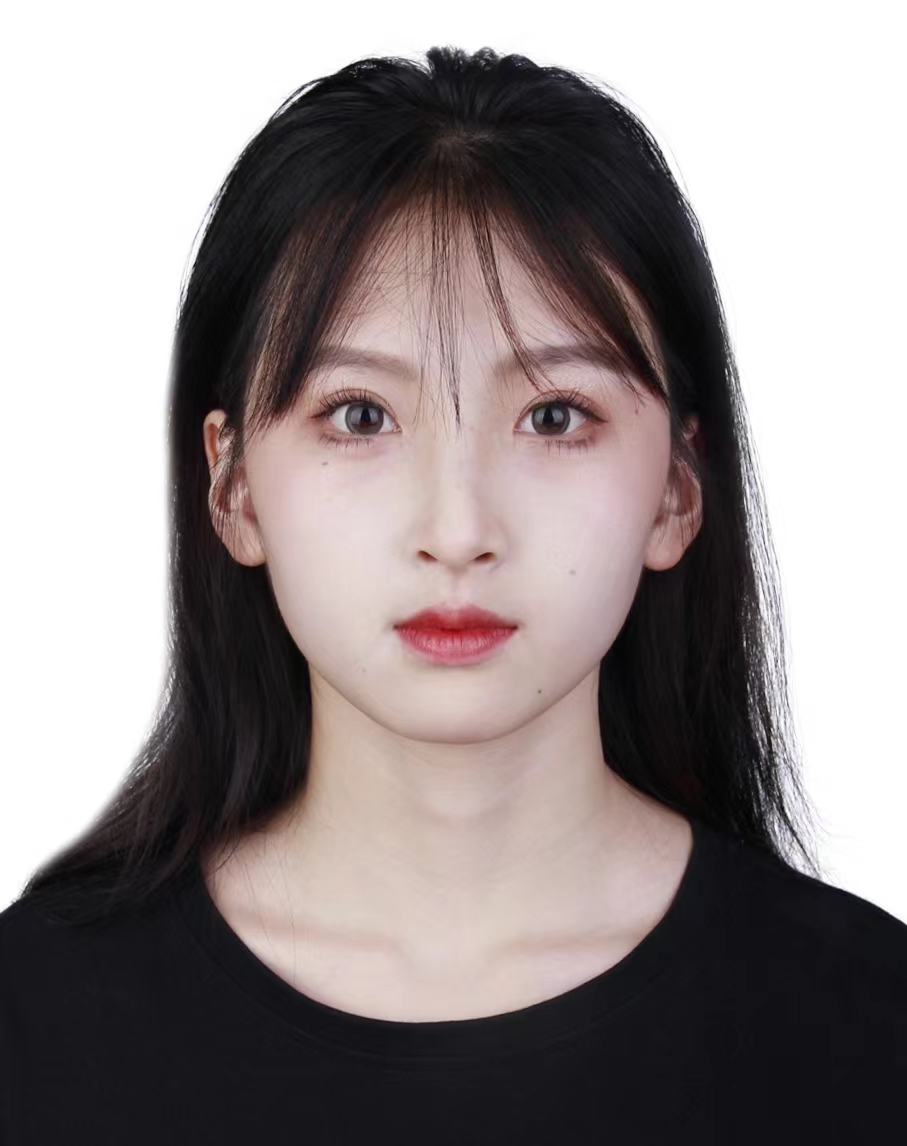}}]{Jiarui Zhao} received her Bachelor's degree in Energy and Electronic Devices from Chang'an University in 2023. She is currently pursuing her Master's degree at the University of the Chinese Academy of Sciences. Her research focuses on the fields of micro/nano energy and sensing technology, with particular interests in control systems and human-computer interaction. She is dedicated to exploring advanced technologies for energy harvesting, high-precision sensing, and the integration of intelligent control in next-generation interactive systems.\end{IEEEbiography}

\begin{IEEEbiography}[{\includegraphics[width=1in,height=1.25in,clip,keepaspectratio]{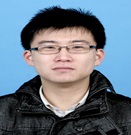}}]{Min Cheng} received her Bachelor of  Science degree in Information Management and InformationSystems from Wannan Medical University in 2010. From 2010 to 2011, I served as an electronic medical record implementation engineer at Winning HEALTH Group Company. From 2011 to 2013, he worked as a smart healthcare project manager at Neusoft Group Company. Since 2013, he has been a senior engineer at the Information Center of Wuhu Hospital, Beijing Anding Hospital, Capital Medical University. His research interests include,but not limited to, medical software engineering methodology, intelligent medicine. At present, he is promoting the planning work for the informatization and intelligence construction of the National Medical Center, and am committed to promoting the deep integration of new-generation information technology and medical services.\end{IEEEbiography}

\begin{IEEEbiography}[{\includegraphics[width=1in,height=1.25in,clip,keepaspectratio]{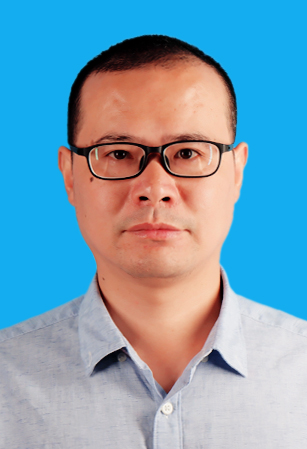}}]{Li Yang} received his Bachelor of
Engineering in Computer Science and Technology from Anhui University in 2006 and his Master of Engineering
in Computer Technology from Anhui University in 2012. In 2019, as a young backbone teacher at Wannan Medical University, he participated in an exchange program at the University of Hong Kong. From 2023 to 2024, he is working as a visiting scholar at the University of Science and Technology of China, focusing on research related to artificial intelligence. Currently, Li Yang is an associate professor at Wannan Medical University. His research interests include but are not limited to artificial intelligence, medical image processing, and the psychology of college students.\end{IEEEbiography}

\begin{IEEEbiography}[{\includegraphics[width=1in,height=1.25in,clip,keepaspectratio]{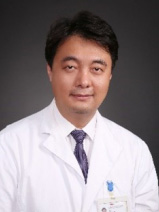}}]{Lei Tian}
is an Associate Professor, Deputy Chief Physician, and Master's Supervisor. He was selected as a Beijing Science and Technology Rising Star, a “Qingmiao” Talent by the Beijing Hospital Authority, a “Distinguished Talent” of Beijing Tongren Hospital, and a “Star of Tomorrow” in the China Optometry Talent Development Program. He serves as a committee member of the Ocular Surface and Dry Eye Section of the Chinese Medical Doctor Association, the National Orthokeratology Safety Monitoring Committee, and the Ophthalmology Branch of the Chinese Medical Equipment Association. He is also the Vice Chairman of the Youth Committee of the Ophthalmology Branch of the Beijing Medical Association. His research interests include ophthalmology, intelligent medicine and medical informatics.
\end{IEEEbiography}

\begin{IEEEbiography}[{\includegraphics[width=1in,height=1.25in,clip,keepaspectratio]{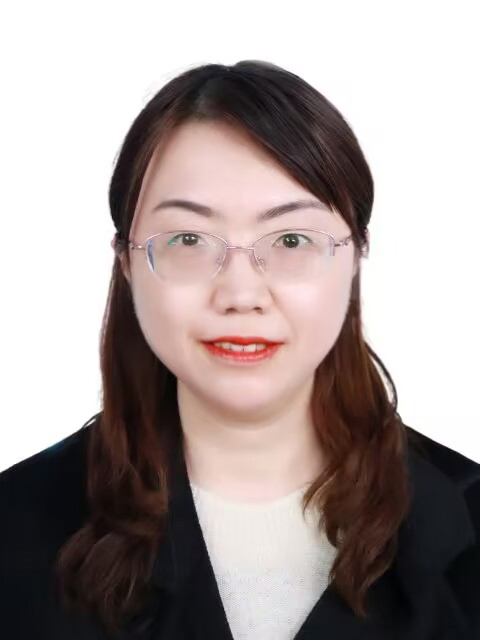}}]{Nan Wan} received her Bachelor of Engineering in Computer Science and Technology from Anhui Normal University in 2003 and her Master's degree in Educational Technology from Anhui Normal University in 2011. She is now an associate professor at Wannan Medical university, with a main research focus on machine learning and intelligent medicine. \end{IEEEbiography}


\end{document}